\documentclass[twocolumn]{article}

\usepackage{arxiv}

\usepackage[utf8]{inputenc} 
\usepackage[T1]{fontenc}    
\usepackage{xcolor}  
\usepackage{hyperref}	
\usepackage{url}            
\usepackage{booktabs}       
\usepackage{amsfonts}       
\usepackage{amsmath}
\usepackage{nicefrac}       
\usepackage{microtype}      
\usepackage[capitalize]{cleveref}       
\usepackage{lipsum}         
\usepackage{graphicx}
\usepackage[numbers]{natbib}
\usepackage{doi}
\usepackage{capt-of}

\usepackage{subcaption}
\usepackage{multirow}
\usepackage{colortbl, siunitx}
\usepackage{pifont}
\usepackage{arydshln}
\usepackage{listings}
\usepackage{xspace}
\usepackage{enumitem}
\usepackage[ruled,linesnumbered]{algorithm2e}

\definecolor{faintgray}{gray}{0.60}   
\definecolor{ourrow}{HTML}{e9f9fb}    
\definecolor{mt}{HTML}{198b9a}
\definecolor{ablrow}{gray}{0.85}

\newcommand{\fainttext}[1]{\textcolor{faintgray}{#1}}

\newcommand\cmark{\ding{51}}   
\newcommand{\xmark}{\ding{55}}  

\makeatletter
\DeclareRobustCommand\onedot{\futurelet\@let@token\@onedot}
\def\@onedot{\ifx\@let@token.\else.\null\fi\xspace}
\makeatother
\def\eg{\emph{e.g}\onedot} 
\def\ie{\emph{i.e}\onedot}

\crefname{section}{Sec.}{Secs.}
\Crefname{section}{Section}{Sections}
\Crefname{table}{Table}{Tables}
\crefname{table}{Tab.}{Tabs.}

\title{BoundMatch: Boundary detection applied to semi-supervised segmentation}

\date{September 30, 2025}

\author{
  Haruya Ishikawa \quad Yoshimitsu Aoki \\
	Department of Electrical Engineering\\
	Keio University\\
	Yokohama, Japan \\
	\texttt{haruyaishikawa@keio.jp}\\
}

\renewcommand{\shorttitle}{\textit{arXiv} Template}

\hypersetup{
  pdftitle={BoundMatch: Boundary detection applied to semi-supervised segmentation},
  pdfsubject={},
  pdfauthor={Haruya Ishikawa, Yoshimitsu Aoki},
  pdfkeywords={Semantic Segmentation, Boundary Detection, Semi-supervised Learning},
}

\begin{document}
\twocolumn[{
  \begin{@twocolumnfalse}

	\maketitle

	\begin{abstract}

Semi-supervised semantic segmentation (SS-SS) aims to mitigate the heavy annotation burden of dense pixel labeling by leveraging abundant unlabeled images alongside a small labeled set.
While current consistency regularization methods achieve strong results, most do not explicitly model boundaries as a separate learning objective.
In this paper, we propose BoundMatch, a novel multi-task SS‑SS framework that explicitly integrates semantic boundary detection into a teacher-student consistency regularization pipeline.
Our core mechanism, Boundary Consistency Regularized Multi-Task Learning (BCRM), enforces prediction agreement between teacher and student models on both segmentation masks and detailed semantic boundaries, providing complementary supervision from two independent tasks.
To further enhance performance and encourage sharper boundaries, BoundMatch incorporates two lightweight fusion modules: Boundary-Semantic Fusion (BSF) injects learned boundary cues into the segmentation decoder, while Spatial Gradient Fusion (SGF) refines boundary predictions using mask gradients, yielding more reliable boundary pseudo-labels.
This framework is built upon SAMTH, a strong teacher-student baseline featuring a Harmonious Batch Normalization (HBN) update strategy for improved stability.
Extensive experiments on diverse datasets including Cityscapes and Pascal VOC show that BoundMatch achieves competitive performance against current state-of-the-art methods.
Our approach achieves state-of-the-art results on the new Cityscapes benchmark with DINOv2 foundation model.
Ablation studies highlight BoundMatch's ability to improve boundary-specific evaluation metrics, its effectiveness in realistic large-scale unlabeled data scenario, and applicability to lightweight architectures for mobile deployment.

	\end{abstract}

  \end{@twocolumnfalse}
  \vspace{1em}
}]


\section{Introduction}
\label{sec:introduction}

Semantic segmentation, the task of assigning a class label to every pixel, provides the detailed scene understanding crucial for many modern vision systems.
It is particularly vital for autonomous driving, where precise identification of road boundaries, pedestrians, and vehicles is paramount for safe navigation \cite{cordts2016cityscapes, Elhassan2024RealtimeSS}.
Beyond driving, it empowers robotic agents to interact intelligently with complex environments \cite{xia2018gibson, narasimhan2020seeing}, aids in clinical diagnostics through medical image analysis \cite{Huang2021GraphBAS3NetBS, Jiao2022LearningWL}, and enables large-scale environmental monitoring via remote sensing \cite{Huang2024DeepLearningBasedSS}.
However, the reliance of high-performance models on vast amounts of dense, pixel-level annotations presents a major bottleneck, as manual labeling is extremely time-consuming and costly.
To address this issue, semi‑supervised semantic segmentation (SS‑SS) offers a compelling solution by aiming to achieve strong performance while drastically reducing annotation requirements, by primarily leveraging large quantities of readily available unlabeled data.

Among various SS-SS strategies, Consistency Regularization (CR) has proven highly effective and is widely adopted, often implemented within teacher-student frameworks \cite{french2020cutmixmt, Yang2022UniMatch}.
In this setup, a ``teacher'' model (\eg an exponential moving average (EMA) of the ``student'') generates pseudo-labels on weakly perturbed unlabeled images, which then supervise the student model trained on strongly perturbed versions of the same images.
This encourages the model to produce consistent predictions despite input perturbations.
While CR methods have significantly boosted overall segmentation accuracy (\eg mIoU), they typically focus on pixel-wise classification without explicit boundary modeling.
Notably, recent fully supervised methods have shown that incorporating boundary detection as an auxiliary multi-task learning (MTL) improves segmentation quality, particularly at object boundaries \cite{takikawa2019gscnn, ishikawa2023SBCB, liao2024mobileseed}.
These boundary-aware approaches help delineate objects more precisely---critical for applications like autonomous driving where clear object separation affects safety.

\begin{figure}[t]
  \centering
  \includegraphics[width=0.8\linewidth]{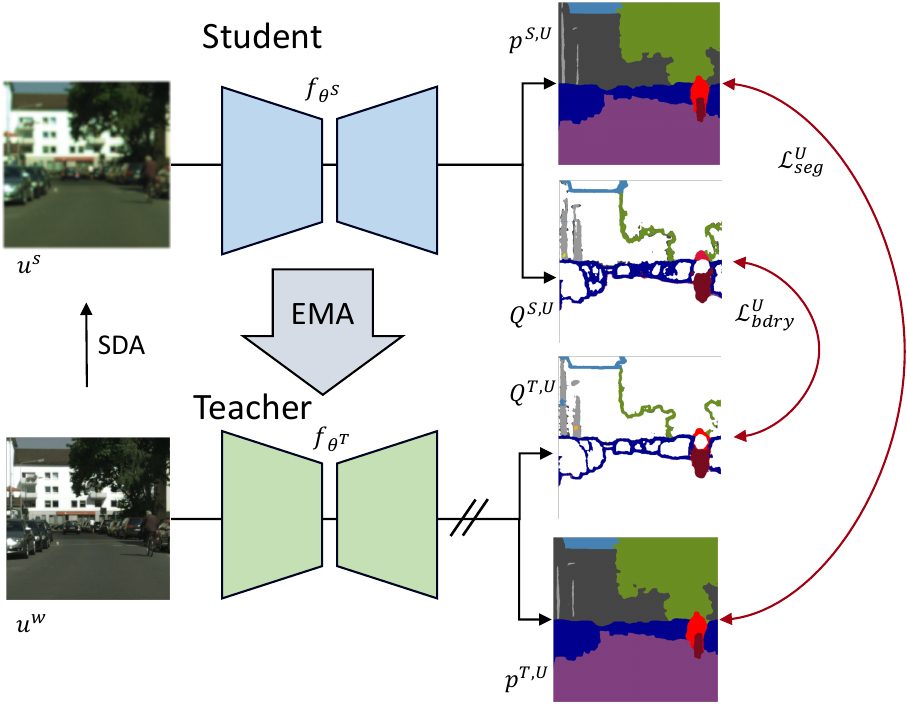}
  \vspace{-7pt} 
  \caption{BoundMatch applies consistency regularization (CR) to both segmentation masks and boundaries in semi-supervised semantic segmentation (SS-SS). Its core mechanism, Boundary Consistency Regularized Multi-Task Learning (BCRM), enforces this by ``matching'' the student's boundary predictions derived from strongly augmented inputs against boundary pseudo-labels generated by the teacher from weakly augmented inputs. To further enhance performance, lightweight fusion modules refine the teacher's initial segmentation and boundary predictions, leading to higher-quality pseudo-labels.}
  \label{fig:method_bcrm}
\vspace{-15pt} 
\end{figure}

While some SS-SS methods have begun exploring boundary information, existing approaches typically fall into two categories: (1) methods that derive boundary cues from segmentation pseudo-labels to re-weight consistency losses near edges \cite{Li2023CFCG,Tarubinga2025CWBASS}, and (2) approaches like BoundaryMatch \cite{Li2024BoundaryMatch} that add binary boundary detection as an auxiliary task. 
However, both categories have limitations.
The former relies on boundaries derived from potentially noisy pseudo-labels, which can propagate segmentation errors to boundary regions.
The latter, while adding boundary detection as an auxiliary task, still derives boundary pseudo-labels from segmentation outputs via Laplacian operators.
This means both tasks are supervised by the same underlying signal, potentially limiting the benefits of multi-task learning where independent supervision signals could provide complementary information.
This analysis motivates our approach: \emph{learning semantic boundaries from hierarchical features independently of segmentation predictions, while establishing explicit fusion pathways that enable bidirectional information exchange between tasks}.
This design aims to provide more independent supervision while maintaining beneficial interactions between the two tasks.

Building on this insight, we propose \textbf{BoundMatch}, a teacher-student framework that integrates semantic boundary detection with consistency regularization for SS-SS.
At its core, BoundMatch introduces \textbf{Boundary Consistency Regularized Multi-Task Learning (BCRM)}, which enforces prediction agreement between teacher and student models on both segmentation masks and semantic boundaries.
Crucially, our boundary head learns from hierarchical backbone features through independent task heads, rather than deriving boundaries from segmentation outputs, enabling it to provide complementary supervision.
To leverage this complementarity, we incorporate two fusion modules: \textbf{Boundary-Semantic Fusion (BSF)} integrates learned boundary cues into the segmentation decoder, while \textbf{Spatial Gradient Fusion (SGF)} refines boundary predictions using segmentation gradients, yielding cleaner boundary pseudo-labels for consistency regularization.
As a foundation for our experiments, we also develop SAMTH, a strong baseline that addresses training instabilities in teacher-student frameworks through \textbf{Harmonious Batch Normalization (HBN)}, which maintains consistent batch normalization statistics between teacher and student networks.
We demonstrate that BoundMatch's modular design can enhance various existing CR methods by integrating it with multiple baselines (\ie SAMTH, UniMatch, and PrevMatch).

Our experiments demonstrate that BoundMatch consistently improves performance across diverse evaluation settings.
On urban-driving datasets (Cityscapes, BDD100K, and SYNTHIA) under varying label ratios, BoundMatch achieves improvements of 0.4--2.4\% mIoU over strong SAMTH baseline, with state-of-the-art results on the recent DINOv2-based Cityscapes benchmark.
The method also performs well in realistic large-scale scenarios with abundant unlabeled data and generalizes to academic benchmarks (Pascal VOC and ADE20K) and remote-sensing data (LoveDA), suggesting that boundary-aware learning can provide benefits across different scenarios.
Through ablation studies, we analyze each component's contribution and demonstrate that improvements extend beyond mIoU to boundary-specific metrics (BIoU and BF1), providing a more complete picture of segmentation quality.
Furthermore, BoundMatch scales to lightweight architectures (MobileNet-V2, AFFormer), maintaining its benefits while meeting real-time deployment constraints.

Our main contributions are:
\begin{itemize}[leftmargin=1em]
  \item We investigate whether semantic boundaries can provide additional supervision in consistency regularization for SS-SS. Through experiments with BCRM, we show that enforcing consistency on both segmentation and boundary predictions from hierarchical features can provide improvements.

  \item We design BoundMatch with two fusion modules (BSF and SGF) that connect boundary and segmentation tasks. Our ablations show these modules contribute incrementally to performance, with the complete framework further improving both mIoU and boundary metrics.

  \item We identify training instabilities in teacher-student frameworks and propose Harmonious Batch Normalization (HBN) as a solution. While orthogonal to our boundary work, HBN improves several baseline methods and yields SAMTH, which serves as a strong baseline for our experiments.
  
  \item We report boundary-specific metrics (BIoU and BF1) alongside mIoU in our experiments, providing additional perspective on segmentation quality. Our results show that improvements extend to these boundary metrics, though we acknowledge that comprehensive evaluation of all prior methods remains infeasible.

  \item We demonstrate that BoundMatch can be integrated with different baseline methods (UniMatch and PrevMatch) and architectures (ResNet, DINOv2-based transformers, lightweight models), showing consistent though incremental improvements across these variations.
\end{itemize}


\section{Related Work}
\label{sec:related_work}

\begin{table*}[t]
\centering
\caption{Comparison of boundary-aware semi-supervised segmentation methods. Our BoundMatch utilizes semantic boundaries to further discriminate between different classes, while prior methods rely on binary edges. Unlike previous approaches that derive boundaries from segmentation outputs (\ie pseudo-labels), BoundMatch learns boundaries independently using a teacher model. Furthermore, we introduce explicit fusion mechanisms (BSF and SGF) to create bidirectional information flow between boundary and segmentation tasks, rather than treating them as independent objectives. These design choices enable BoundMatch to achieve performance gains, evaluated using both standard mIoU and boundary-specific metrics (BIoU and BF1).}
\label{tab:method_comparison}
\footnotesize
\setlength{\tabcolsep}{6pt} 
\renewcommand{\arraystretch}{1.3} 
\begin{tabular}{lcccc}
\toprule
 & CFCG & CW-BASS & BoundaryMatch & \textbf{BoundMatch (Ours)} \\
 & 2023 \cite{Li2023CFCG} & 2025 \cite{Tarubinga2025CWBASS} & 2024 \cite{Li2024BoundaryMatch} & \\
\midrule
\textbf{Boundary Type}               & Binary & Binary & Binary & \textbf{Semantic} \\
\textbf{Boundary Source}             & Derived (Laplacian) & Derived (Sobel) & Derived (Laplacian) & \textbf{Learned (Teacher pred.)} \\
\textbf{Integration}                 & Loss reweighting & Loss reweighting & Parallel MTL & \textbf{Fused MTL (BSF+SGF)} \\
\textbf{Boundary Consistency Reg.}   & -- & -- & \cmark & \cmark \\
\textbf{Bidirectional Flow}          & -- & -- & \xmark & \cmark \\
\textbf{Feature Stages}              & -- & -- & Single-stage & \textbf{Hierarchical} \\
\bottomrule
\end{tabular}
\vspace{-10pt} 
\end{table*}

\paragraph{Semantic Segmentation and Boundary-Aware Strategies.}
Semantic segmentation has evolved significantly, progressing from early fully convolutional networks (FCNs) \cite{shelhamer2015fcn} to advanced architectures incorporating multi-scale features \cite{zhao2017pspnet}, attention mechanisms \cite{Zhu2019ANN}, and Transformer-based models \cite{xie2021segformer}.
The DeepLabV3+ architecture \cite{chen2018deeplabv3plus}, featuring an encoder-decoder structure and atrous convolutions, remains a popular choice due to its effectiveness in capturing both local and global context.
Real-time semantic segmentation has also garnered attention, leading to efficient architectures like BiSeNet \cite{Yu2020BiSeNetV2} and AFFormer \cite{Dong2023AFFormer}.

Recognizing that segmentation accuracy is often hindered by imprecise object boundaries, boundary-aware methods have emerged.
GSCNN \cite{takikawa2019gscnn} was among the first to achieve significant improvements by incorporating multi-task learning (MTL) of boundaries and segmentation.
DecoupleSegNet \cite{li2020decouple} introduced specialized decoders for boundary and interior regions.
Inspired by dedicated boundary detection methods \cite{xie2015hed,liu2017rcf,yu2017casenet}, RPCNet \cite{zhen2020rpcnet} proposed an MTL framework integrating both tasks.
CSEL \cite{yu2021coupled} employed an affinity learning strategy to further refine segmentation.
More recent work has focused on conditioning the backbone features with boundary information, exemplified by STDC \cite{Fan2021STDC} and SBCB \cite{ishikawa2023SBCB}.
Mobile-Seed \cite{liao2024mobileseed} extended boundary-aware MTL to lightweight Transformer backbones suitable for mobile deployment.
However, the application of MTL with semantic boundary detection within the SS-SS context remains largely unexplored, motivating our work.

\paragraph{Semi-Supervised Semantic Segmentation (SS-SS).}
Fully supervised semantic segmentation methods require large amounts of costly pixel-wise annotations.
SS-SS aims to mitigate this burden by leveraging abundant unlabeled data alongside a smaller labeled set.
Consistency regularization (CR) has become mainstream in SS-SS, aiming to produce more reliable pseudo-labels and to manage noisy pseudo-labels via perturbation agreement \cite{Hu2021AEL,Kwon2022ELN,Wang2023CVCC,Sun2023CorrMatch,Mai2024RankMatch,Wang2024DDFP,Yin2024UCCL}; analyzed by NRCR \cite{Zhang2024NRCR} in terms of label refinement, strong regularization, multi-view learning, robust losses, and sample selection.
A common implementation involves the mean-teacher framework \cite{french2020cutmixmt}, where an exponential moving average (EMA) of the student model's weights generates stable pseudo-labels for unlabeled data.
CutMix-MT \cite{french2020cutmixmt} integrated CutMix augmentation to enhance consistency.
Subsequent works like AugSeg \cite{Zhao2022AugSeg} and iMAS \cite{Zhao2022iMAS} explored stronger augmentation strategies.
Teacher-Student framework has been further extended with various techniques, such as using multiple teachers (\eg DualTeacher \cite{Na2023DualTeacher}) and contrastive learning (\eg ReCo \cite{Liu2021ReCo}, U2PL \cite{Wang2022U2PL}).

There are other notable directions in CR.
Some methods focus on co-training to provide cross supervision by using two or more independent networks with different initializations or architectures (\eg CPS \cite{Chen2021CPS} and Diverse Co-Training \cite{Li2023DiverseCoT}).
On the other hand, approaches like UniMatch \cite{Yang2022UniMatch} and its variants \cite{Shin2024PrevMatch,Ma2025DLMSI} have proposed a unified framework that combines strong-to-weak perturbation regularization with a single model training strategy.

Our work builds upon these CR foundations by introducing boundary detection as an auxiliary task, an aspect largely missing in prior SS-SS literature.
Recently, applying SS-SS principles to Transformer-based architectures has gained traction \cite{Wang2024AllSpark,Hoyer2023SemiVL,yang2025unimatchv2,Than2025SegKC}.
Recognizing the need to validate SS-SS in diverse, practical scenarios, we also evaluate our approach on DPT \cite{Ranftl2021DPT} with DINOv2 pretrained ViT backbones \cite{Oquab2023DINOv2} and lightweight architectures suitable for mobile deployment.

\paragraph{Boundary-Aware SS-SS.}
While most SS-SS methods concentrate on pseudo-label refinement and consistency losses, explicit integration of boundary information remains relatively unexplored.
\cref{tab:method_comparison} summarizes the key differences among existing boundary-aware SS-SS approaches.

Some methods derive boundary masks from segmentation pseudo-labels using edge detection operators: CFCG \cite{Li2023CFCG} applies Laplacian operators while CW-BASS \cite{Tarubinga2025CWBASS} uses Sobel filters to identify boundary regions for loss re-weighting.
These approaches are inherently limited to the boundary information already present in the pseudo-labels.

Other methods explicitly learn boundaries as an auxiliary task. Graph-BAS$^3$Net \cite{Huang2021GraphBAS3NetBS} pioneered this direction for semi-supervised medical image segmentation.
BoundaryMatch \cite{Li2024BoundaryMatch} recently added binary boundary detection to UniMatch through parallel multi-task learning.
However, as shown in \cref{tab:method_comparison}, BoundaryMatch derives its boundary pseudo-labels from segmentation outputs using Laplacian operators, potentially propagating noise from the segmentation predictions.
Additionally, it uses only binary edges and single-stage features, reporting limited gains when integrated with modern SS-SS methods.

Our BoundMatch explores a different combination of design choices:
(1) Semantic boundaries rather than binary edges, providing class-specific information;
(2) Learned boundary predictions from a teacher model rather than deriving them from pseudo-labels;
(3) Bidirectional fusion between tasks (BSF and SGF) rather than parallel multi-task learning;
(4) Hierarchical features from multiple backbone stages \cite{zhen2020rpcnet,ishikawa2023SBCB} for multi-scale boundary reasoning.
Table 1 summarizes the key differences among existing boundary-aware SS-SS approaches.

These design choices enable BoundMatch to achieve performance gains as measured by both standard mIoU and boundary-specific metrics (BIoU and BF1).
Our results suggest that integrating learned semantic boundaries with explicit fusion mechanisms between boundary and segmentation tasks can be more effective than treating them as independent objectives.


\begin{figure}[t]
  \centering
  \includegraphics[width=0.70\linewidth]{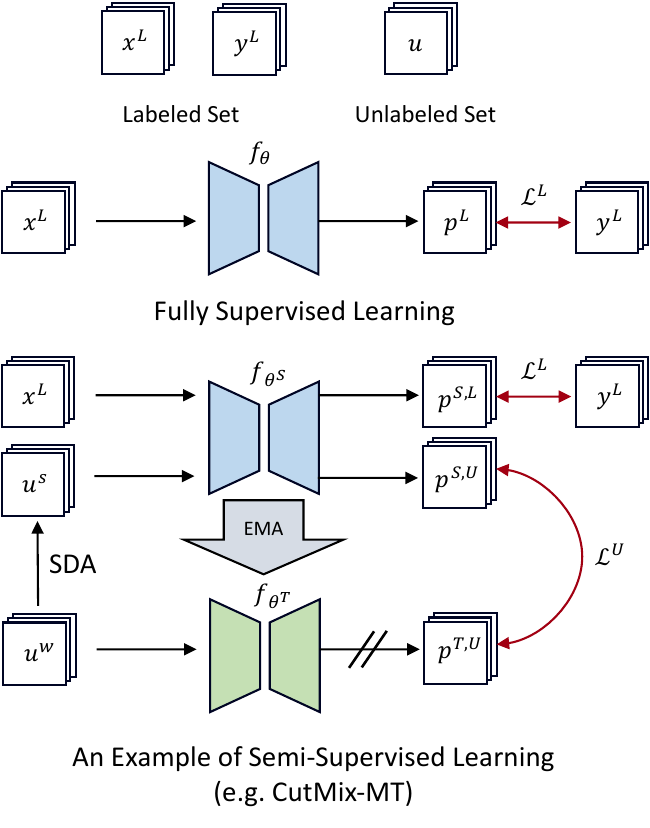}
  \vspace{-10pt} 
  \caption{Illustration of the Teacher-Student framework, a common approach for consistency regularization (CR) in semi-supervised semantic segmentation (SS-SS). This framework utilizes unlabeled images by having the teacher model generate pseudo-labels from weakly augmented inputs ($u^w$). These pseudo-labels then supervise the student model, which processes corresponding strongly augmented inputs ($u^s$), thereby enforcing prediction consistency. The teacher model is typically updated via an Exponential Moving Average (EMA) of the student's weights, and its predictions are detached from the backpropagation graph (indicated by dashed lines).}
  \label{fig:method_notations}
\vspace{-10pt} 
\end{figure}

\section{Approach}
\label{sec:approach}

In this section, we present BoundMatch, our multi-task framework that incorporates boundary detection into semi-supervised semantic segmentation (SS-SS).
Our design integrates multiple noise-handling mechanisms identified in recent SS-SS literature \cite{Zhang2024NRCR}: EMA-based label refinement, strong data augmentation, multi-task learning with independent segmentation and boundary heads, and confidence-based sample selection.
The boundary detection task, learned from hierarchical features rather than derived from segmentation outputs, provides complementary supervision that can be particularly beneficial at object boundaries.

We begin by outlining the notation and preliminaries for SS-SS, including the teacher-student paradigm, in \cref{sec:notation,sec:teacher_student}.
We then introduce SAMTH, our baseline method detailed in \cref{sec:samth}.
Building upon this baseline, \cref{sec:bcrm} describes our core Boundary Consistency Regularized Multi-Task Learning (BCRM) framework.
Subsequently, \cref{sec:refine} details the complementary boundary-aware modeling and refinement strategies: Boundary-Semantic Fusion (BSF) and Spatial Gradient Fusion (SGF).
Finally, \cref{sec:boundmatch} synthesizes these components (SAMTH, BCRM, BSF, and SGF) into the complete BoundMatch framework.

\subsection{Notation and Preliminaries}
\label{sec:notation}

Let $\mathcal{D}^L = \{(x_i, y_i)\}_{i=1}^{N}$ be the labeled dataset, where \(x_i\) denotes the input image and \(y_i \in \{0, 1, \dots, C-1\}^{H \times W}\) represents the corresponding semantic segmentation map with $C$ semantic classes and spatial dimensions $H \times W$.
In addition, let $\mathcal{D}^U = \{u_j\}_{j=1}^{M}$ be an unlabeled dataset consisting of images $u_j$.

The goal of SS-SS is to learn a segmentation model $f_\theta(\cdot)$ parameterized by $\theta$ by leveraging both $\mathcal{D}^L$ and $\mathcal{D}^U$.
The overall loss is formulated as:
\begin{equation}
\mathcal{L} = \mathcal{L}^L + \lambda\, \mathcal{L}^U,
\end{equation}
where $\mathcal{L}^L$ is the supervised loss over the labeled set and $\mathcal{L}^U$ is the unsupervised loss applied to the unlabeled set.
The pixel-wise cross-entropy loss $H$ where we can represent the overall supervised loss as:
\begin{equation}
\mathcal{L}^L = \frac{1}{B^L}\sum_{b=1}^{B^L}\frac{1}{HW} \sum_{i,j} H\Big(y^{L}(b,i,j), p^{S,L}(b,i,j)\Big),
\end{equation}
with $B^L$ representing mini-batch size and \(p^{L} = f_{\theta}(x)\) representing the semantic segmentation prediction for the network.

\subsection{Teacher-Student Framework for Consistency Regularization}
\label{sec:teacher_student}

The teacher-student framework, as shown in \cref{fig:method_notations}, is a popular paradigm for SS-SS.
It employs consistency regularization (CR) between the predictions of a teacher network and a student network for unlabeled samples.
The core idea is to encourage the student network to produce consistent predictions for an unlabeled image under different perturbations.
The teacher network, which is typically an exponential moving average (EMA) of the student network, provides more stable and reliable predictions, which are then used to guide the student's learning.

In our framework, the teacher's output from weakly augmented images is used to generate hard pseudo-labels that supervise the student's predictions under strong data augmentation (SDA).
This process can be seen as a form of self-distillation, where the student learns from the teacher's predictions on unlabeled data.

Specifically, given a weakly augmented image \(u^w\) and its strongly augmented counterpart \(u^s\), we first obtain the teacher's prediction \(p^{T,w}\) and generate hard pseudo-labels $\hat{p}^T = \operatorname{argmax}(p^{T,w})$.
These pseudo-labels represent the teacher's most confident predictions for each pixel in the unlabeled image.

The unsupervised consistency loss is then defined as:

\scriptsize
\begin{equation}\begin{split}
  \mathcal{L}^U &= R(p^{S,s}, p^{T,w}) = \frac{1}{B^U} \sum_{b=1}^{B^U}\frac{1}{HW}\sum_{i,j}\mathbf{1}\Big(\max\big(p^{T,w}(b,i,j)\big) \geq \tau\Big) \\
  &\qquad H\Big(\hat{p}^T(b,i,j), p^{S,s}(b,i,j)\Big),
\end{split}\end{equation}
\normalsize
where \(\mathbf{1}(\cdot)\) is an indicator function, and \(\tau\) is the confidence threshold to filter out noisy pseudo-labels.
This loss function measures the discrepancy between the student's predictions on the strongly augmented unlabeled image and the teacher's hard pseudo-labels on the weakly augmented version.
By minimizing this loss, the student is encouraged to learn from the teacher's predictions and become more robust to different augmentations.

The student's parameters are jointly updated using labeled set via $\mathcal{L}^L$.
The teacher’s parameters \(\theta^T\) are updated using EMA of the student’s parameters:
\begin{equation}
\theta^T \leftarrow \alpha\, \theta^T + (1 - \alpha)\, \theta^S,
\end{equation}
with decay rate \(\alpha\).
EMA helps to stabilize the teacher's predictions by averaging the student's parameters over time, leading to more reliable pseudo-labels.

\subsection{Strong Data Augmentation Mean-Teacher with Hard Pseudo-Labels (SAMTH)}
\label{sec:samth}

Our baseline, SAMTH, builds upon the CutMix-MT framework \cite{french2020cutmixmt} but distinguishes itself by using thresholded pseudo-labels (hard pseudo-labels) rather than Softmax probabilities (soft pseudo-labels), similar to recent methods such as AugSeg \cite{Zhao2022AugSeg}.

A key differentiator of SAMTH is its \textbf{Harmonious Batch Normalization (HBN)} update strategy, addressing potential instabilities in teacher-student training using Batch Normalization (BN) with EMA.
Common alternatives risk issues:
\begin{itemize}[leftmargin=1em]
  \item Forwarding only partial data through the teacher (\eg CutMix-MT \cite{french2020cutmixmt}, U2PL \cite{Wang2022U2PL}) can lead to biased BN statistics unsuitable for the teacher's full role.
  \item Copying or taking the EMA of the student's BN statistics (\eg AugSeg \cite{Zhao2022AugSeg}, iMAS \cite{Zhao2022iMAS}) couples the teacher's normalization to the student's potentially noisy state and risks incompatibility between these borrowed BN statistics and the teacher's own EMA-updated weights.
\end{itemize}

HBN aims to mitigate these concerns.
We forward the complete batch ($x, u^w, u^s$) through both networks: $p^{S,L}, p^{S,w}, p^{S,s} = f_{\theta^S}(x, u^w, u^s)$ and $p^{T,L}, p^{T,w}, p^{T,s} = f_{\theta^T}(x, u^w, u^s)$.
Crucially, the teacher's BN layers remain in \lstinline{train} mode, updating statistics independently based on the full data stream passing through the teacher's own parameters.
For every iteration, the teacher updates its BN statistics, while the rest of the model parameters (\ie layer weights) are EMA-updated from the student's parameters.
This ensures the teacher's BN statistics are \textit{harmonious} with its specific weights and the full input distribution, decoupling its normalization from student fluctuations.
This promotes more stable training dynamics and improves performance, as validated empirically across multiple baselines in our ablations (\cref{sec:effect_HBN}).
For more complete details and pseudocode, please refer to Appendix~\ref{app:hbn}.

\subsection{Boundary Consistency Regularized Multi-Task Learning (BCRM)}
\label{sec:bcrm}

\begin{figure*}[t]
  \centering
  \includegraphics[width=0.90\textwidth]{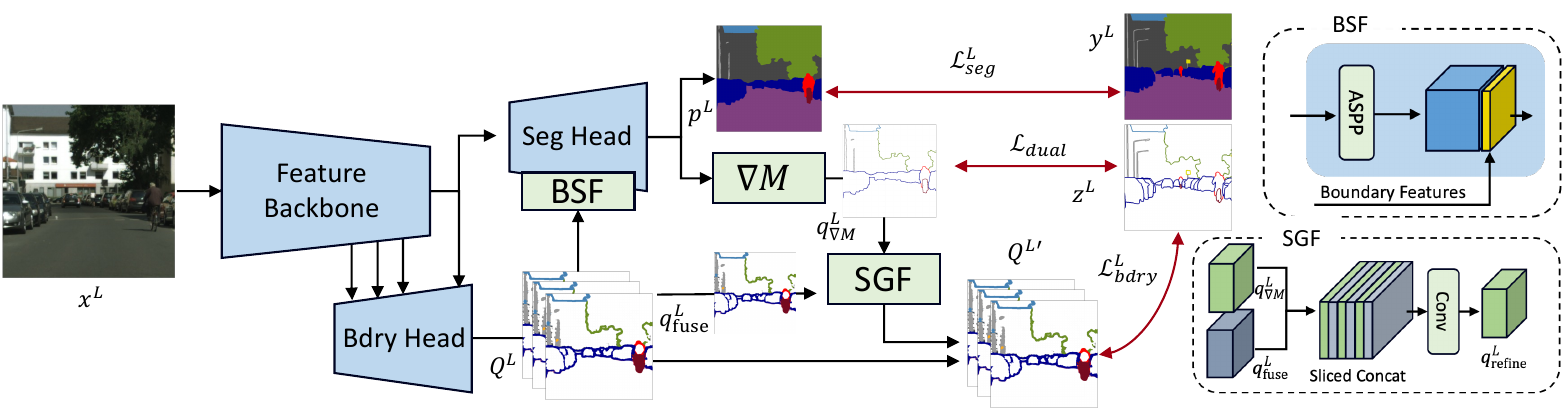}
  \vspace{-5pt} 
  \caption{Overview of the BoundMatch architecture processing labeled samples, illustrating multi-task learning (MTL) of boundary detection alongside segmentation. The Boundary-Semantic Fusion (BSF) module integrates learned boundary cues from the boundary head into the segmentation head's features, promoting sharper segmentation boundaries. Concurrently, the Spatial Gradient Fusion (SGF) module refines the boundary predictions by fusing them with the spatial gradient of the segmentation mask ($\nabla M$). This refinement improves the quality of boundary supervision for the labeled loss and is important for generating cleaner boundary pseudo-labels used in the consistency regularization process for unlabeled data.}
  \label{fig:method_refine}
\vspace{-10pt} 
\end{figure*}

While many semantic segmentation approaches rely solely on segmentation supervision, recent works have demonstrated the benefit of multi-task learning (MTL) with a boundary detection task.
BCRM takes this concept further by applying consistency regularization (CR) to both segmentation and boundary predictions.

\subsubsection{Notation and Boundary Detection Head}
\label{sec:boundary_head}

We adopt a lightweight boundary detection head inspired by SBCB \cite{ishikawa2023SBCB} and CASENet \cite{yu2017casenet}.
The architecture leverages hierarchical features from the backbone (\eg features from Stem, layer1, layer2, and layer4 of a ResNet).
Features from the earlier layers are processed to yield single-channel outputs, while the layer4 feature produces an output with \(C\) channels, where \(C\) is the number of classes.
Specifically, the features from these layers are passed through a single $1 \times 1$ convolutional layer, respectively, each followed by BN and ReLU activation.
These processed features are then bilinearly upsampled to $\nicefrac{1}{2}$ of the input image size.
Following \cite{ishikawa2023SBCB}, we use an additional $3\times3$ convolutional layer to reduce artifacts from upsampling.
The outputs are then fused using a sliced concatenation and undergo a grouped convolution to form the final boundary prediction map of shape \(C \times H \times W\).
For more details on the boundary detection head architecture, please refer to Appendix~\ref{app:boundary_head}.

Let \(Q\) denote the set of boundary predictions (\eg fused output \(q_{\text{fuse}}\) and the layer4 output \(q_{\text{last}}\)).
For the boundary detection task, the ground-truth boundary is denoted by \(z\) and the boundary prediction \(q\) is a sigmoid-activated probability map in \([0,1]\).
We supervise these outputs using a reweighted pixel-wise binary cross-entropy loss:
\begin{equation}
H_{bdry}(z, q) = \beta\, (1-z)\, \log(1 - q) + (1-\beta)\, z\, \log(q),
\end{equation}
where the weights are defined by \(\beta = \nicefrac{|z^+|}{|z|} \), \(\quad 1-\beta = \nicefrac{|z^-|}{|z|}\), with \(|z^+|\) and \(|z^-|\) representing the number of boundary and non-boundary pixels, respectively.
This reweighting addresses the class imbalance between boundary and non-boundary pixels, ensuring that the model learns to detect boundaries.

For semantic (multi-label) boundaries, we use a reweighted multi-label cross-entropy loss over the \(C\) channels.
The supervised boundary loss is then given by:

\scriptsize
\begin{equation}
  \mathcal{L}_{bdry}^L = \sum_{q^{S,L}\in Q^L}\frac{1}{B^L}\sum_{b=1}^{B^L}\frac{1}{C H W} \sum_{c,i,j} H_{bdry}\Big(z(b,c,i,j),\, q^{S,L}(b,c,i,j)\Big).
\end{equation}
\normalsize

While our primary auxiliary task is semantic boundary detection, we also explore the use of binary boundaries in \cref{sec:binary_vs_semantic}.
We opt for semantic boundaries because they provide richer, class-specific edge information compared to generic binary edges.
This detailed supervision can encourage the model to learn features that better distinguish between adjacent classes, potentially leading to improved feature representations in the backbone \cite{ishikawa2023SBCB}.

\subsubsection{MTL for SS-SS with Boundaries}

For the labeled set, the boundary detection head is trained alongside the segmentation head.
For unlabeled data, we perform consistency regularization analogous to segmentation as shown in \cref{fig:method_bcrm}.
We generate boundary pseudo-labels from the teacher's weakly augmented predictions $q^{T,w}$, where $q^{T,w} = q_{\text{fuse}}^{T,w}$ represents the teacher's fused boundary output (as described in \cref{sec:boundary_head}).
These are converted to hard pseudo-labels using threshold $\tau_{bdry}$: $\hat{q}^T = \mathbf{1}\Big(q^{T,w} \geq \tau_{bdry}\Big)$.
The unsupervised boundary loss then enforces consistency between these pseudo-labels and the student's strongly augmented predictions:

\scriptsize
\begin{equation}
  \mathcal{L}_{bdry}^U \\= \sum_{q^{S,s}\in Q^U}\frac{1}{B^U} \sum_{b=1}^{B^U}\frac{1}{C H W}\sum_{c,i,j}H_{bdry}\Big(\hat{q}^T(b,c,i,j),\, q^{S,s}(b,c,i,j)\Big).
\end{equation}
\normalsize

Note that when SGF is enabled (\cref{sec:sgf}), we use the refined boundary prediction $q_{\text{refine}}^{T,w}$ instead of $q_{\text{fuse}}^{T,w}$ to generate higher-quality pseudo-labels, as the refinement process produces cleaner boundaries by incorporating spatial gradient information.

From a noise-robustness perspective, BCRM leverages multi-view learning principles where the segmentation and boundary detection heads serve as \textit{two complementary views} of the same visual understanding task.
Unlike methods that derive boundaries directly from segmentation outputs (\ie BoundaryMatch \cite{Li2024BoundaryMatch} creates dependent pseudo-labels), our boundary head learns from hierarchical backbone features independently, providing a distinct supervisory signal.
This independence is important: when one task branch generates erroneous pseudo-labels, the other branch can continue learning from its own independently-generated pseudo-labels, preventing error propagation between tasks.
Following noise-robust learning principles identified in \cite{Zhang2024NRCR}, such multi-view consistency with independent learning objectives aims to helps prevent overfitting to task-specific noise patterns.

\subsection{Boundary-Aware Modeling and Refinement}
\label{sec:refine}

To further enhance the boundary quality and boost segmentation performance, we propose two complementary modules: \textbf{Boundary-Semantic Fusion (BSF)} and \textbf{Spatial Gradient Fusion (SGF)}.
BSF integrates learned boundary cues into the segmentation decoder to improve object delineation, while SGF refines boundary predictions using spatial gradients from the segmentation mask to produce cleaner boundary pseudo-labels.
These simple fusion mechanisms establish bidirectional information flow between tasks as shown in \cref{fig:method_refine}.

\subsubsection{Boundary-Semantic Fusion (BSF)}
\label{sec:bsf}

Inspired by GSCNN \cite{takikawa2019gscnn}, BSF explicitly integrates boundary predictions into the segmentation head through concatenation-based fusion. 
Formally, let $F_\text{ASPP}\in \mathbf{R}^{(C_\text{ASPP} \times H/16 \times W/16)}$ denote the ASPP features and $q\in \mathbf{R}^{(C\times H/16 \times W/16)}$ denote the downsampled boundary features from the boundary prediction head which is detached from gradient computation.
The BSF operation is defined as:
\begin{equation}
  F_\text{fused} = \text{Conv}_{1\times 1}([F_\text{ASPP}; q])
\end{equation}
where $[\cdot;\cdot]$ denotes concatenation along the channel dimension, and $\text{Conv}_{1\times 1}$ is the existing bottleneck layer in the DeepLabV3+ decoder.
This design choice preserves full information from both modalities without information loss through element-wise operations.
We emphasize that BSF introduces no additional layers; we only expand the input channels of the existing bottleneck layer from $C_\text{ASPP}$ to $C_\text{ASPP} + C$, maintaining architectural simplicity while enabling effective feature fusion.
By incorporating boundary cues, BSF aims to provide the segmentation decoder with complementary edge information that may assist in object delineation.

\subsubsection{Spatial Gradient Fusion (SGF)}
\label{sec:sgf}

\begin{figure}[t]
  \centering
  \includegraphics[width=0.8\linewidth]{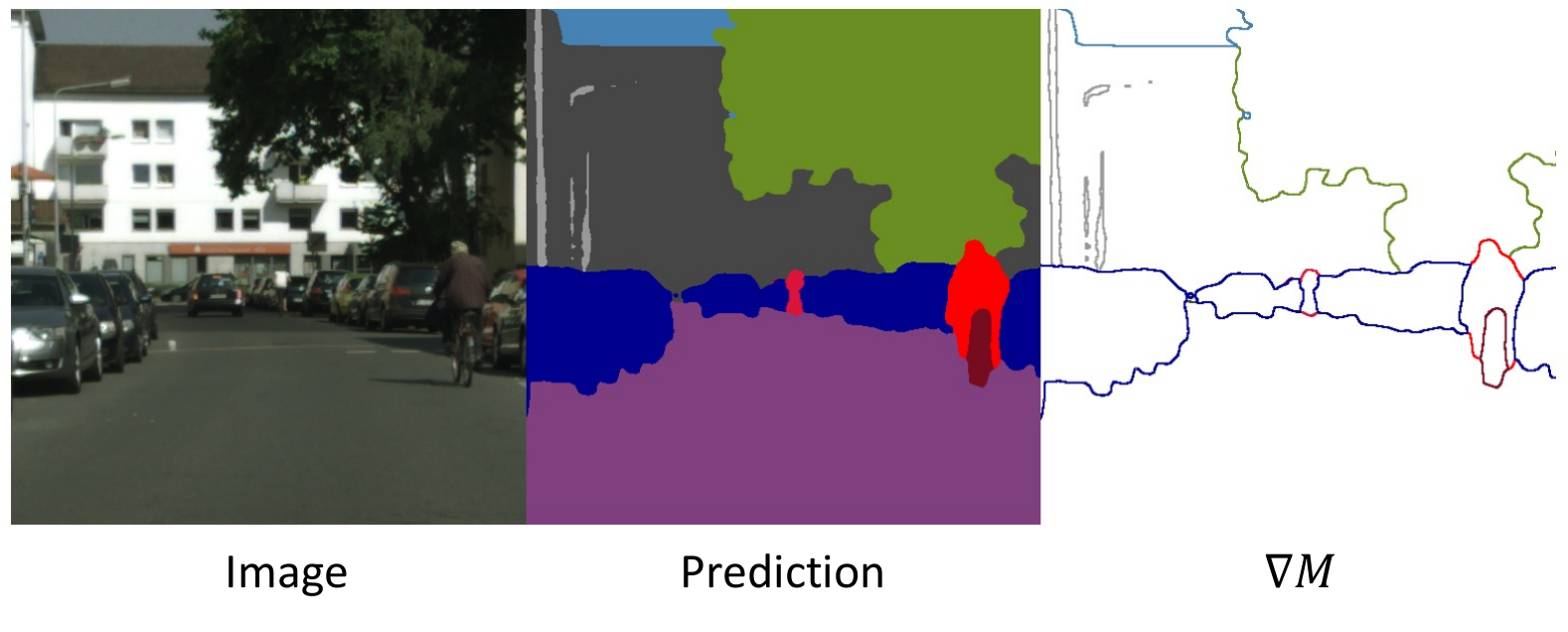}
  \vspace{-10pt} 
  \caption{Visualization of the output of spatial gradient operator on segmentation prediction.}
  \label{fig:method_sgf}
\vspace{-10pt} 
\end{figure}

While the boundary detection head learns from hierarchical features, it can benefit from geometric cues present in the segmentation predictions.
We adopt the spatial gradient operator from RPCNet \cite{zhen2020rpcnet} to extract these cues.
The spatial gradient of the semantic mask \(M\) is computed as:
\begin{equation}
\nabla M(i,j) = \left\| M(i,j) - \operatorname{pool}\big(M(i,j)\big) \right\|,
\end{equation}
where \(\operatorname{pool}\) denotes adaptive average pooling with a \(3 \times 3\) kernel.
An example of the spatial gradient operator is shown in \cref{fig:method_sgf}.
The resulting gradient map, $q_{\nabla M}$, is concatenated with the boundary prediction $q_\text{fuse}$ for each channel:
\begin{equation}
  [q_{\text{fuse},0},\, q_{\nabla M,0},\, q_{\text{fuse},1},\, q_{\nabla M,1},\, \dots,\, q_{\text{fuse},C-1},\, q_{\nabla M,C-1}],
\end{equation}
forming a \(2C\)-channel activation map.
This interleaving arrangement ensures that each paired boundary-gradient features for the same class are processed together.
The map is then processed by a single \(C\)-grouped convolutional layer:
\begin{equation}
q_\text{refine} = \text{GroupConv}_C([q_\text{fuse,0}, q_{\nabla M,0}, \dots])
\end{equation}
where $\text{GroupConv}_C$ denotes a grouped convolution with $C$-groups.
This design choice follows established semantic boundary detection methods \cite{yu2017casenet,hu2019dff} where per-class grouped convolutions enable class-specific boundary refinement.
Each group independently refines its corresponding class boundary using the paired spatial gradient information, preserving class-specific edge patterns while maintaining computational efficiency through a single convolution operation.
The refined boundaries are supervised using the same \(H_{bdry}\) loss with the labeled set.

Moreover, we introduce a duality loss on the labeled set to enforce consistency between the spatial gradient and the ground-truth boundaries:
\begin{equation}
  \mathcal{L}_{dual} = \frac{1}{B^L}\sum_{b=1}^{B^L}\frac{1}{C H W}\sum_{c,i,j} \left| q^L_{\nabla M}(b,c,i,j) - z(b,c,i,j) \right|.
\end{equation}
We apply this loss only on labeled samples because the ground-truth boundary annotations are reliable; applying it on unlabeled data, which relies on pseudo-labels, introduces instability.

\subsection{BoundMatch}
\label{sec:boundmatch}

Finally, BCRM combined with BSF and SGF forms the \textbf{BoundMatch} framework.
The labeled and unlabeled losses when we apply BoundMatch to SAMTH (SAMTH+BoundMatch) is
\begin{equation}
\mathcal{L}^{L} = \mathcal{L}^L_{seg} + \lambda_{bdry} \,\mathcal{L}^L_{bdry} + \mathcal{L}_{dual},
\end{equation}
\begin{equation}
\mathcal{L}^{U} = \lambda_{seg}\,\mathcal{L}^U_{seg} + \lambda_{bdry}\,\mathcal{L}^U_{bdry},
\end{equation}
and the total loss is
\begin{equation}
\mathcal{L} = \mathcal{L}^{L} + \lambda\, \mathcal{L}^{U}.
\end{equation}

The design of BoundMatch emphasizes modularity and flexibility.
The framework's core principles---boundary CR through independent task heads, bidirectional fusion between tasks, and hierarchical feature extraction---are architecture-agnostic.
This modularity enables practitioners to adopt different configurations based on their requirements: full BoundMatch for maximum accuracy, BCRM+SGF for maintaining inference speed, or BCRM alone for minimal overhead.
While demonstrated primarily with SAMTH, BoundMatch integrates naturally with other CR methods.
For UniMatch \cite{Yang2022UniMatch}, we incorporate boundary consistency alongside its existing feature-level and strong-view losses, using an EMA teacher for stable boundary pseudo-labels.
For PrevMatch \cite{Shin2024PrevMatch}, we leverage its robust ensemble pseudo-labels for boundary consistency, circumventing single-iteration instabilities.

BoundMatch's approach differs from prior methods by learning boundaries independently through dedicated task heads rather than deriving them from segmentation outputs.
The teacher model generates boundary pseudo-labels from its learned boundary head (potentially refined by SGF), providing supervision that is complementary to, rather than dependent on, segmentation predictions.
This independence aims to provide noise robustness: when segmentation pseudo-labels are uncertain at object edges, the boundary task can provide additional supervision signals, and vice versa.
The implementation of BSF and SGF relies on standard operations---feature concatenation and grouped convolutions---which facilitates integration with various segmentation architectures without requiring substantial modifications.

\section{Experiments}
\label{sec:experiments}

In this section, we validate our proposed approach, BoundMatch, through comprehensive experiments.
We first benchmark BoundMatch against recent state-of-the-art SS-SS methods on several benchmark datasets.
Subsequently, we present detailed ablation studies to analyze the individual contributions of our core components (BCRM, BSF, SGF), evaluate the effect of the chosen boundary formulation (semantic vs. binary), and assess the impact of our Harmonious Batch Normalization (HBN) update strategy.
Finally, we report the computational cost associated with BoundMatch and applications to more real-world settings.

\subsection{Experimental Configuration}
\label{sec:experimental_configuration}

Here, we specify the experimental configuration used for evaluating BoundMatch.
This includes the datasets utilized (covering urban scenes and academic benchmarks), the model architectures, and key implementation details.
We follow the standard SS-SS evaluation protocols for each dataset, ensuring a fair comparison with existing methods following \cite{Yang2022UniMatch}.

\subsubsection{Datasets}
\label{sec:datasets}

We primarily evaluate our method on urban scene datasets, including Cityscapes, BDD100K, and SYNTHIA, while also considering Pascal VOC 2012 and ADE20K to assess performance in more diverse scenarios.
In all semi-supervised learning protocols, the images not selected as labeled data form the unlabeled set for training.

\noindent\textbf{Cityscapes dataset} \cite{cordts2016cityscapes} is tailored for semantic understanding of urban street scenes.
It comprises 2,975 high-resolution training images and 500 validation images, with annotations for 19 semantic categories.
In our experiments, we follow the splits defined by UniMatch \cite{Yang2022UniMatch} and evaluate on label partitions corresponding to $\nicefrac{1}{16}$, $\nicefrac{1}{8}$, and $\nicefrac{1}{4}$ of the full labeled training set.

\noindent\textbf{BDD100K dataset} \cite{Yu2018BDD100K}, originally designed for multi-task learning in autonomous driving, contains 100,000 video frames.
For our segmentation experiments, we utilize a subset of 8,000 images (\(1280 \times 720\) resolution), divided into 7,000 for training and 1,000 for validation.
As BDD100K has not been widely used previously for SS-SS evaluation, we define labeled splits corresponding to $\nicefrac{1}{64}$, $\nicefrac{1}{32}$, and $\nicefrac{1}{16}$ of the available training annotations.

\noindent\textbf{SYNTHIA dataset} \cite{Ros2016Synthia} provides synthetic urban scenes generated via computer graphics. We utilize the SYNTHIA-RAND ("Rand") subset, containing 13,400 images (\(1280 \times 760\) resolution) with annotations aligned to the Cityscapes semantic categories.
For our experiments, we use a split of 7,000 training images and 1,000 validation images, evaluating on labeled partitions corresponding to $\nicefrac{1}{64}$, $\nicefrac{1}{32}$, and $\nicefrac{1}{16}$.

\noindent\textbf{PASCAL VOC 2012} \cite{pascal-voc-2012} is a standard object recognition benchmark consisting of 10,582 images annotated for 21 classes (20 object categories plus background).
Following established SS-SS protocols, we considered two training settings: the ``Classic'' protocol uses only the high-quality subset of 1,464 images for the labeled pool, while the ``Blender'' protocol samples labeled images randomly from the entire dataset.

\noindent\textbf{ADE20K dataset} \cite{Zhou2017ADE20k} offers diverse scenes with dense annotations covering 150 semantic categories.
It includes 20,210 training images and 2,000 validation images.
Following prior work, we evaluate on label partitions of $\nicefrac{1}{128}$, $\nicefrac{1}{64}$, and $\nicefrac{1}{32}$ of the training set.

\subsubsection{Implementation Details}
\label{sec:implementation_details}

Our experiments utilize ResNet-50 and ResNet-101 as encoder backbones across all datasets, coupled with a DeepLabV3+ segmentation head using an output stride of 16 for training efficiency.
For generating semantic and binary boundary labels, we employ the on-the-fly (OTF) strategy from \cite{ishikawa2023SBCB} by applying signed distance function and using distance of 2 to obtain per-class boundaries.
This approach integrates boundary extraction into the data loading pipeline, guaranteeing consistent boundary widths despite random resizing augmentations.

Key hyperparameters are configured per dataset group:

\noindent\textbf{Cityscapes, BDD100K, and SYNTHIA:} We use a base learning rate of $0.01$.
Loss weights are set to \(\lambda_{seg} = 1.0\) and \(\lambda_{bdry} = 1.0\), with an overall unsupervised loss weight \(\lambda = 1.0\).
The EMA decay is \(\alpha = 0.99\), and the confidence threshold for segmentation pseudo-labels is \(\tau=0\).
\(\tau_{bdry}\) is set to 0.5.
Training uses a mini-batch size of 16 (8 labeled, 8 unlabeled) for 80,000 iterations.
The input crop size is \(801 \times 801\) for Cityscapes and \(641 \times 641\) for BDD100K and SYNTHIA.

\noindent\textbf{Pascal VOC:} Hyperparameters are adjusted: base learning rate is $0.001$ and boundary loss weight \(\lambda_{bdry}=0.1\) (other weights remain \(\lambda_{seg}=1.0, \lambda=1.0\)).
EMA decay is increased to \(\alpha = 0.999\), and the segmentation confidence threshold is raised to \(\tau = 0.95\).
\(\tau_{bdry}\) is set to 0.5.
We use a mini-batch size of 32 (16 labeled, 16 unlabeled) and train for 120,000 iterations with a \(321 \times 321\) crop size.

\noindent\textbf{ADE20K:} We adopt the same loss weights (\(\lambda_{seg}, \lambda_{bdry}, \lambda\)), EMA decay (\(\alpha\)), and thresholds (\(\tau\) and \(\tau_{bdry}\)) as Pascal VOC.
However, specific training parameters differ: the base learning rate is $0.01$, mini-batch size is 16 (8 labeled, 8 unlabeled), training duration is 80,000 iterations, and the input crop size is \(513 \times 513\).

Training employs the stochastic gradient descent (SGD) optimizer with a polynomial learning rate decay scheduler.
Image augmentations follow UniMatch practices \cite{Yang2022UniMatch}, combining weak transformations (resizing, random cropping, horizontal flipping) with strong augmentations (color jittering, grayscale conversion, Gaussian blurring, CutMix).
To stabilize the unsupervised loss components, we apply a sigmoid ramp-up schedule: \(\lambda_{seg}\) increases from 0 to its target value over the first 15\% of iterations, while \(\lambda_{bdry}\) ramps up linearly throughout the entire training process.

Mean intersection over union (mIoU) is the primary evaluation metric for semantic segmentation, but is known to be insensitive to boundary-specific errors \cite{Csurka2013WhatIA, Li2020AutoSegLoss}.
To quantitatively assess boundary improvements, we adopt \textbf{Boundary IoU (BIoU)} and \textbf{Boundary F1 Score (BF1)}, which explicitly measure boundary alignment and detection quality, respectively (see Appendix~\ref{app:boundary_metrics} for detailed definitions).

All experiments are implemented using PyTorch and the \texttt{mmsegmentation} library \cite{mmseg2020}.
Training is conducted on dual-GPU setups, using either two NVIDIA RTX 3090 or two NVIDIA A6000 GPUs, depending on the memory footprint of the specific benchmark and model configuration.

We use the same experimental setup for all our ablation studies, in \cref{sec:ablation} unless otherwise specified.

\begin{figure*}[t]
  \centering
  \includegraphics[width=0.90\linewidth]{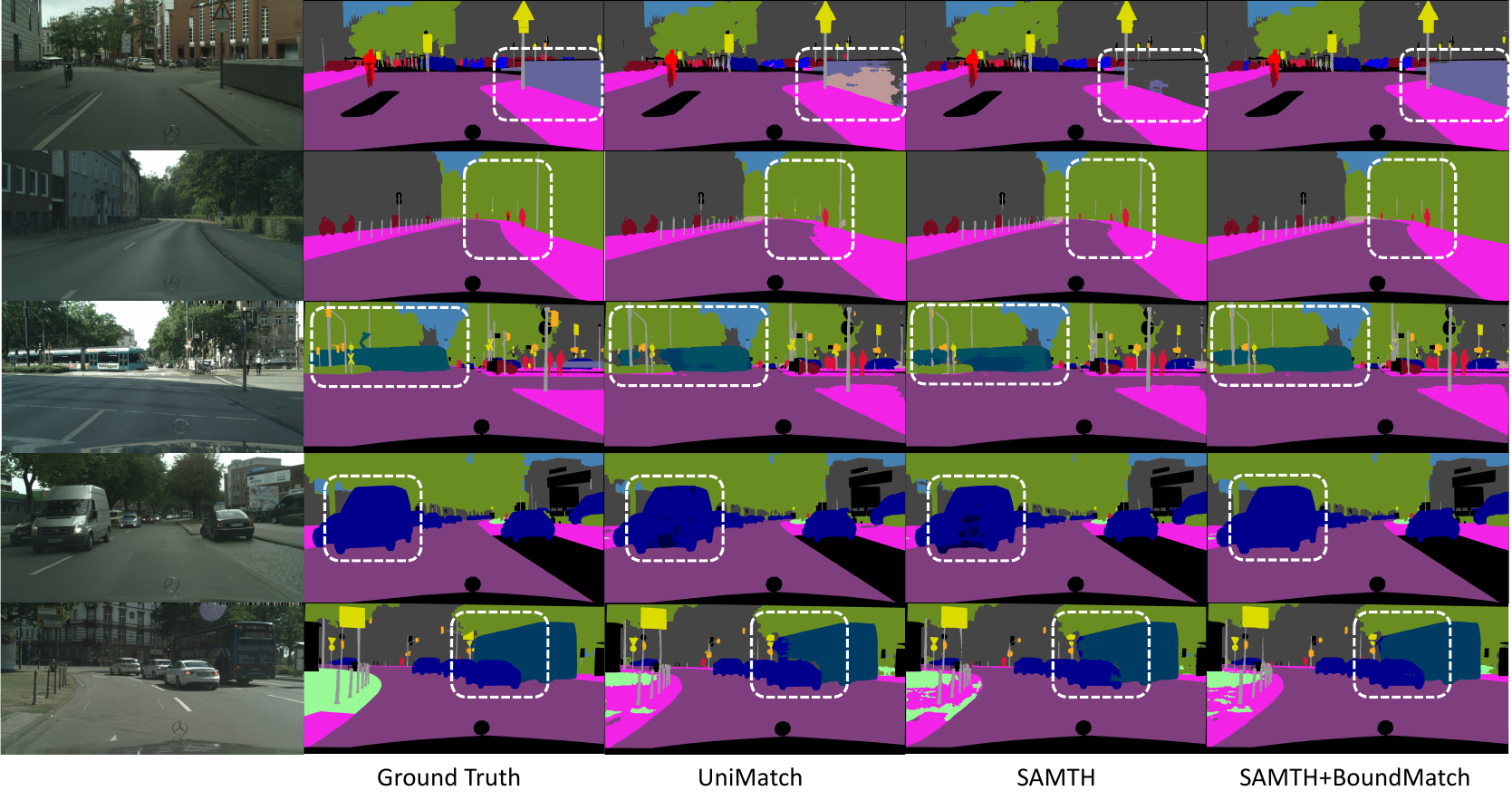}
  \vspace{-5pt} 
  \caption{Qualitative results on the Cityscapes dataset on $\nicefrac{1}{16}$ split. We compare UniMatch \cite{Yang2022UniMatch} with our SAMTH baseline and SAMTH+BoundMatch. Our method produces less segmentation errors especially for object boundaries and difficult categories. Best viewed zoomed in and in color.}
  \label{fig:results_qualitative}
\vspace{-10pt} 
\end{figure*}

\begin{table}[t]
\centering
\caption{Comparison with recent state-of-the-art methods on Cityscapes.  
All methods use DeepLabV3$+$ with ResNet-50/101 backbones.
$\dag$ denotes method reproduced by us using the same training partition protocol.
\textbf{Bold} and \underline{underline} denotes the best and the second-best results, respectively.
Results are averaged over three runs following \cite{Yang2022UniMatch}.}
\label{tab:benchmark_cityscapes}
\scriptsize
\setlength{\tabcolsep}{3.5pt}
\begin{tabular}{
  l l
  S S S  
  S S S  
}
\toprule
\multicolumn{2}{c}{} &
\multicolumn{3}{c}{\textbf{ResNet-50}} &
\multicolumn{3}{c}{\textbf{ResNet-101}}\\
\cmidrule(lr){3-5}\cmidrule(l){6-8}
\multicolumn{2}{c}{} &
{$\nicefrac{1}{16}$} & {$\nicefrac{1}{8}$} & {$\nicefrac{1}{4}$} &
{$\nicefrac{1}{16}$} & {$\nicefrac{1}{8}$} & {$\nicefrac{1}{4}$}\\
\midrule
Supervised &  & 63.3 & 70.2 & 73.1 & 66.3 & 72.8 & 75.0 \\

U2PL \cite{Wang2022U2PL} & \fainttext{[CVPR'22]} &
70.6 & 73.0 & 76.3 & 70.3 & 74.4 & 76.5 \\

CVCC \cite{Wang2023CVCC} & \fainttext{[CVPR'23]} &
74.9 & 76.4 & 77.3 & {-}  & {-}  & {-}  \\

iMAS \cite{Zhao2022iMAS} & \fainttext{[CVPR'23]} &
74.3 & 77.4 & 78.1 & {-}  & {-}  & {-}  \\

AugSeg \cite{Zhao2022AugSeg} & \fainttext{[CVPR'23]} &
73.7 & 76.5 & 78.8 & 75.2 & 77.8 & 79.6 \\

CFCG$^\dag$ \cite{Li2023CFCG} & \fainttext{[ICCV'23]} &
75.0 & 76.9 & 78.8 & 76.8 & 78.4 & 79.5 \\

LogicDiag \cite{Liang2023LogicDiag} & \fainttext{[ICCV'23]} &
{-} & {-} & {-} &
76.8 & \underline{78.9} & \textbf{80.2} \\

CSS \cite{Wang2023CSS} & \fainttext{[ICCV'23]} &
{-} & {-} & {-} & 74.0 & 76.9 & 77.9 \\

Co-T. \cite{Li2023DiverseCoT} & \fainttext{[ICCV'23]} &
{-} & 76.3 & 77.1 & 75.0 & 77.3 & 78.7 \\

DAW \cite{Sun2023DAW} & \fainttext{[NeurIPS'23]} &
75.2 & 77.5 & 79.1 & 76.6 & 78.4 & 79.4 \\

CorrMatch \cite{Sun2023CorrMatch} & \fainttext{[CVPR'24]} &
{-} & {-} & {-} & 77.3 & 78.5 & 79.4 \\

RankMatch \cite{Mai2024RankMatch} & \fainttext{[CVPR'24]} &
75.4 & 77.7 & \underline{79.2} & 77.1 & 78.6 & 80.0 \\

DDFP \cite{Wang2024DDFP} & \fainttext{[CVPR'24]} &
{-} & {-} & {-} & 77.1 & 78.2 & 79.9 \\

BoundaryMatch \cite{Li2024BoundaryMatch} & \fainttext{[IS'24]} &
{-} & {-} & {-} & 76.0 & 78.4 & 79.1 \\

UCCL \cite{Yin2024UCCL} & \fainttext{[ICASSP'25]} &
75.8 & 77.2 & 78.2 & {-} & {-} & {-} \\

NRCR \cite{Zhang2024NRCR} & \fainttext{[NN'25]} &
{-} & 77.0 & 77.9 & {-} & 78.2 & 79.5 \\

DMSI \cite{Ma2025DLMSI} & \fainttext{[TMM'25]} &
75.9 & \underline{78.0} & \underline{79.2} & 77.0 & \textbf{79.0} & 79.8 \\

CW-BASS \cite{Tarubinga2025CWBASS} & \fainttext{[ArXiv'25]} &
75.0 & 77.2 & 78.4 & {-} & {-} & {-} \\

\midrule

\rowcolor{ourrow}
\multicolumn{2}{l}{\textbf{SAMTH (ours)}} &
75.1 & 77.3 & 78.9 & 75.5 & 77.9 & 79.7 \\
\rowcolor{ourrow}
\multicolumn{2}{l}{\textbf{SAMTH + BoundMatch (ours)}} &
\textbf{76.5} & \textbf{78.1} & \textbf{79.3} &
\textbf{77.9} & \textbf{79.0} & \underline{80.1} \\
\rowcolor{ourrow}
              & & \textcolor{mt}{+1.4} & \textcolor{mt}{+0.8} & \textcolor{mt}{+0.4} &
              \textcolor{mt}{+2.4} & \textcolor{mt}{+1.1} & \textcolor{mt}{+0.4} \\

\midrule

UniMatch$^\dag$ \cite{Yang2022UniMatch} & \fainttext{[CVPR'23]} &
75.2 & 76.9 & 77.5 & 76.6 & 77.8 & 79.1 \\
\rowcolor{ourrow}
\multicolumn{2}{l}{\textbf{UniMatch + BoundMatch (ours)}} &
76.0 & 77.6 & 78.2 & 77.4 & 78.4 & 79.7 \\
\rowcolor{ourrow}
     & & \textcolor{mt}{+0.8} & \textcolor{mt}{+0.7} & \textcolor{mt}{+0.7} &
     \textcolor{mt}{+0.8} & \textcolor{mt}{+0.6} & \textcolor{mt}{+0.6} \\

\midrule

PrevMatch$^\dag$ \cite{Shin2024PrevMatch} & \fainttext{[ArXiv'24]} &
75.7 & 77.6 & 78.6 & 77.4 & \underline{78.9} & 79.9 \\
\rowcolor{ourrow}
\multicolumn{2}{l}{\textbf{PrevMatch + BoundMatch (ours)}} &
\underline{76.3} & \underline{78.0} & 79.1 &
\underline{77.8} & \textbf{79.0} & \textbf{80.2} \\
\rowcolor{ourrow}
                 & &  \textcolor{mt}{+0.6} & \textcolor{mt}{+0.4} & \textcolor{mt}{+0.5} &
                 \textcolor{mt}{+0.4} & \textcolor{mt}{+0.1} & \textcolor{mt}{+0.3} \\

\bottomrule
\end{tabular}
\vspace{-5pt} 
\end{table}

\begin{table}[t]
\centering
\caption{Comparison of \textbf{SAMTH + BoundMatch} with UniMatch on three datasets using DeepLabV3$+$ (ResNet-50).}
\label{tab:benchmark_etc}
\scriptsize
\setlength{\tabcolsep}{2.8pt}

\begin{tabular}{
  l  
  *{3}{S}  
  *{3}{S}  
  *{3}{S}  
}
\toprule
\multirow{2}{*}{\textbf{Method}} &
\multicolumn{3}{c}{\textbf{BDD100K}} &
\multicolumn{3}{c}{\textbf{SYNTHIA}} &
\multicolumn{3}{c}{\textbf{ADE20K}}\\
\cmidrule(lr){2-4}\cmidrule(lr){5-7}\cmidrule(l){8-10}
& {\small$\nicefrac{1}{64}$} & {\small$\nicefrac{1}{32}$} & {\small$\nicefrac{1}{16}$} &
  {\small$\nicefrac{1}{64}$} & {\small$\nicefrac{1}{32}$} & {\small$\nicefrac{1}{16}$} &
  {\small$\nicefrac{1}{128}$} & {\small$\nicefrac{1}{64}$} & {\small$\nicefrac{1}{32}$}\\
\midrule
Supervised &
40.4 & 45.3 & 52.1 &
63.3 & 68.5 & 69.7 &
7.2  & 9.9  & 13.7 \\

UniMatch$^\dag$ \cite{Yang2022UniMatch} &
49.2 & 52.3 & 56.4 &
68.5 & 71.9 & 72.4 &
13.6 & 18.3 & 23.9 \\

\rowcolor{ourrow}
\textbf{Ours} &
\textbf{52.4} & \textbf{53.9} & \textbf{57.8} &
\textbf{69.9} & \textbf{73.0} & \textbf{74.8} &
\textbf{15.6} & \textbf{19.6} & \textbf{25.3} \\
\bottomrule
\end{tabular}
\vspace{-5pt} 
\end{table}

\begin{table}[t]
\centering
\caption{Comparison with recent state-of-the-art methods on the Pascal VOC 2012 dataset using the \textit{Classic} splits.
All methods are trained using DeepLabV3$+$ (ResNet-50/101).}
\label{tab:benchmark_voc_classic}
\scriptsize
\setlength{\tabcolsep}{3.6pt}

\begin{tabular}{
  l l
  *{3}{S}  
  *{3}{S}  
}
\toprule
\multicolumn{2}{c}{\multirow{2}{*}{\textbf{Classic}}} &
\multicolumn{3}{c}{\textbf{ResNet-50}} &
\multicolumn{3}{c}{\textbf{ResNet-101}} \\
\cmidrule(lr){3-5}\cmidrule(l){6-8}
\multicolumn{2}{c}{} &
{92} & {183} & {366} & {92} & {183} & {366} \\
\midrule
Supervised & &
44.0 & 52.3 & 61.7 &
45.1 & 55.3 & 64.8 \\

CVCC \cite{Wang2023CVCC} & \fainttext{[CVPR'23]} &
{-} & {-} & {-} & 70.2 & 74.4 & 77.4 \\

iMAS \cite{Zhao2022iMAS} & \fainttext{[CVPR'23]} &
{-} & {-} & {-} & 68.8 & 74.4 & 78.5 \\

AugSeg \cite{Zhao2022AugSeg} & \fainttext{[CVPR'23]} &
64.2 & 72.2 & 76.2 & 71.1 & 75.5 & 78.8 \\

ESL \cite{Ma2023ESL} & \fainttext{[ICCV'23]} &
{-} & 69.5 & 72.6 & 71.0 & 74.1 & 78.1 \\

CSS \cite{Wang2023CSS} & \fainttext{[ICCV'23]} &
68.0 & 71.9 & 74.9 & {-} & {-} & {-} \\

Co-T. \cite{Li2023DiverseCoT} & \fainttext{[ICCV'23]} &
73.1 & 74.7 & 77.1 & 75.7 & 77.7 & \underline{80.1} \\

DAW \cite{Sun2023DAW} & \fainttext{[NeurIPS'23]} &
68.5 & 73.1 & 76.3 & 74.8 & 77.4 & 79.5 \\

CorrMatch \cite{Sun2023CorrMatch} & \fainttext{[CVPR'24]} &
{-} & {-} & {-} & 76.4 & 78.5 & 79.4 \\

RankMatch \cite{Mai2024RankMatch} & \fainttext{[CVPR'24]} &
71.6 & 74.6 & 76.7 & 75.5 & 77.6 & 79.8 \\

BoundaryMatch \cite{Li2024BoundaryMatch} & \fainttext{[IS'24]} &
{-} & {-} & {-} & 75.4 & 77.3 & 79.3 \\

UCCL \cite{Yin2024UCCL} & \fainttext{[ICASSP'25]} &
{-} & 74.1 & 77.1 & {-} & {-} & {-} \\

NRCR \cite{Zhang2024NRCR} & \fainttext{[NN'25]} &
{-} & {-} & {-} & \underline{77.4} & \textbf{79.7} & \textbf{80.2} \\

DMSI \cite{Ma2025DLMSI} & \fainttext{[TMM'25]} &
{-} & {-} & {-} & 76.5 & 77.4 & 79.7 \\

CW-BASS \cite{Tarubinga2025CWBASS} & \fainttext{[ArXiv'25]} &
72.8 & \textbf{75.8} & 76.2 & {-} & {-} & {-} \\

\midrule

\rowcolor{ourrow}
\textbf{SAMTH (ours)} & &
70.7 & 72.1 & 76.1 &
73.2 & 76.4 & 78.5 \\

\rowcolor{ourrow}
\multicolumn{2}{l}{\textbf{SAMTH + BoundMatch (ours)}} &
72.6 & 73.8 & 77.3 &
76.6 & 78.3 & 78.9 \\
\rowcolor{ourrow}
     & & \textcolor{mt}{+1.9} & \textcolor{mt}{+1.7} & \textcolor{mt}{+1.2} &
     \textcolor{mt}{+3.4} & \textcolor{mt}{+1.9} & \textcolor{mt}{+0.4} \\

\midrule

UniMatch$^\dag$ \cite{Yang2022UniMatch} & \fainttext{[CVPR'23]} &
71.9 & 72.5 & 76.0 &
75.2 & 77.2 & 78.8 \\

\rowcolor{ourrow}
\multicolumn{2}{l}{\textbf{UniMatch + BoundMatch (ours)}} &
\underline{74.2} & \textbf{75.8} & 76.9 &
76.0 & 78.2 & 78.9 \\
\rowcolor{ourrow}
     & & \textcolor{mt}{+2.3} & \textcolor{mt}{+3.3} & \textcolor{mt}{+0.9} &
     \textcolor{mt}{+0.8} & \textcolor{mt}{+1.0} & \textcolor{mt}{+0.1} \\

\midrule

PrevMatch$^\dag$ \cite{Shin2024PrevMatch} & \fainttext{[ArXiv'24]} &
73.4 & \underline{75.4} & \underline{77.5} &
77.0 & 78.5 & 79.6 \\

\rowcolor{ourrow}
\multicolumn{2}{l}{\textbf{PrevMatch + BoundMatch (ours)}} &
\textbf{74.5} & \textbf{75.8} & \textbf{77.7} &
\textbf{77.5} & \underline{78.7} & 79.7 \\

\rowcolor{ourrow}
              & &  \textcolor{mt}{+1.1} & \textcolor{mt}{+0.4} & \textcolor{mt}{+0.2} &
              \textcolor{mt}{+0.5} & \textcolor{mt}{+0.2} & \textcolor{mt}{+0.1} \\

\bottomrule
\end{tabular}
\vspace{-10pt} 
\end{table}

\begin{table}[t]
\centering
\caption{Comparison with recent state-of-the-art methods on the Pascal VOC 2012 dataset using the \textit{Blender} splits. All methods are trained using DeepLabV3$+$ (ResNet-50).}
\label{tab:benchmark_voc_blender}
\footnotesize
\setlength{\tabcolsep}{3.8pt}

\begin{tabular}{
  l l
  *{3}{S}  
}
\toprule
\multicolumn{2}{c}{\textbf{Blender}} &
{\small$\nicefrac{1}{16}$} & {\small$\nicefrac{1}{8}$} & {\small$\nicefrac{1}{4}$}\\
\midrule
Supervised & & 62.4 & 68.2 & 72.3 \\

CVCC \cite{Wang2023CVCC} & \fainttext{[CVPR'23]} &
74.5 & 76.1 & 76.4 \\

iMAS \cite{Zhao2022iMAS} & \fainttext{[CVPR'23]} &
74.8 & 76.5 & 77.0 \\

AugSeg \cite{Zhao2022AugSeg} & \fainttext{[CVPR'23]} &
74.7 & 76.0 & 77.2 \\

CFCG$^\dag$ \cite{Li2023CFCG} & \fainttext{[ICCV'23]} &
75.2 & 76.7 & 77.1 \\

DAW \cite{Sun2023DAW} & \fainttext{[NeurIPS'23]} &
76.2 & 77.6 & 77.4 \\

RankMatch \cite{Mai2024RankMatch} & \fainttext{[CVPR'24]} &
\underline{76.6} & 77.8 & 78.3 \\

IpxMatch \cite{Wu2024IpxMatch} & \fainttext{[IJCNN'24]} &
74.5 & 74.9 & 75.0 \\

NRCR \cite{Zhang2024NRCR} & \fainttext{[NN'25]} &
\underline{76.6} & \textbf{78.2} & 78.7 \\

DMSI \cite{Ma2025DLMSI} & \fainttext{[TMM'25]} &
76.3 & 76.9 & 77.2 \\

\midrule

\rowcolor{ourrow}
\textbf{SAMTH (ours)} & &
73.5 & 75.8 & 76.9 \\

\rowcolor{ourrow}
\multicolumn{2}{l}{\textbf{SAMTH + BoundMatch (ours)}} &
76.3 & 77.2 & 78.1 \\
\rowcolor{ourrow}
     & & \textcolor{mt}{+2.8} & \textcolor{mt}{+1.4} & \textcolor{mt}{+1.2} \\

\midrule

UniMatch$^\dag$ \cite{Yang2022UniMatch} & \fainttext{[CVPR'23]} &
76.0 & 76.9 & 76.7 \\

\rowcolor{ourrow}
\multicolumn{2}{l}{\textbf{UniMatch + BoundMatch (ours)}} &
\underline{76.6} & \underline{77.9} & \textbf{78.9} \\
\rowcolor{ourrow}
                 & & \textcolor{mt}{+0.6} & \textcolor{mt}{+1.0} & \textcolor{mt}{+2.2} \\

\midrule

PrevMatch$^\dag$ \cite{Shin2024PrevMatch} & \fainttext{[ArXiv'24]} &
75.8 & 77.0 & 77.7 \\

\rowcolor{ourrow}
\multicolumn{2}{l}{\textbf{PrevMatch + BoundMatch (ours)}} &
\textbf{76.7} & 77.8 & \underline{78.8} \\
\rowcolor{ourrow}
              & & \textcolor{mt}{+0.9} & \textcolor{mt}{+0.8} & \textcolor{mt}{+1.1} \\

\bottomrule
\end{tabular}
\vspace{-10pt} 
\end{table}

\begin{table}[t]
\centering
\caption{Comparison with recent state-of-the-art methods using DPT with DINOv2 backbones on the Cityscapes dataset.}
\label{tab:benchmark_dinov2}
\footnotesize
\setlength{\tabcolsep}{3.8pt}

\begin{tabular}{
  l l
  *{2}{S}  
  *{2}{S}  
}
\toprule
\multicolumn{2}{c}{\multirow{2}{*}{\textbf{Methods}}} &
\multicolumn{2}{c}{\textbf{DINOv2-S}} &
\multicolumn{2}{c}{\textbf{DINOv2-B}} \\
\cmidrule(lr){3-4}\cmidrule(l){5-6}
\multicolumn{2}{c}{} &
{\small$\nicefrac{1}{16}$} & {\small$\nicefrac{1}{8}$} &
{\small$\nicefrac{1}{16}$} & {\small$\nicefrac{1}{8}$} \\
\midrule
Supervised & & 77.2 & 80.2 & 80.8 & 82.7 \\

UniMatch-V2 \cite{yang2025unimatchv2} & \fainttext{[TPAMI'25]} &
80.6 & 81.9 & 83.6 & 84.3 \\

SegKC \cite{Than2025SegKC} & \fainttext{[CoRR'25]} &
81.2 & 82.4 & {--} & {--} \\

\midrule

\rowcolor{ourrow}
\textbf{SAMTH (ours)} & &
80.8 & 82.0 & 83.2 & 84.1 \\

\rowcolor{ourrow}
\multicolumn{2}{l}{\textbf{SAMTH + BoundMatch (ours)}} &
\textbf{81.5} & \textbf{82.9} & \textbf{84.0} & \textbf{84.8} \\

\rowcolor{ourrow}
              & & \textcolor{mt}{+0.7} & \textcolor{mt}{+0.9} & \textcolor{mt}{+0.8} & \textcolor{mt}{+0.7} \\

\bottomrule
\end{tabular}
\vspace{-10pt} 
\end{table}

\subsection{Comparisons with State-of-the-Art Methods}
\label{sec:benchmarks}

\subsubsection{Cityscapes}
\label{sec:benchmark_cityscapes}

\cref{tab:benchmark_cityscapes} summarizes the quantitative results on the Cityscapes dataset across various labeled data splits ($\nicefrac{1}{16}$, $\nicefrac{1}{8}$, $\nicefrac{1}{4}$) for both ResNet-50 and ResNet-101 backbones.
We have also included boundary metrics in \cref{tab:cityscapes_boundary_metrics} in the Appendix.
Compared with recent state-of-the-art methods, SAMTH+BoundMatch achieves competitive performances, outperforming most evaluation protocols.
Notably, incorporating the boundary-aware components allows SAMTH+BoundMatch to consistently outperform the SAMTH baseline and achieves 1.4\% and 2.4\% improvements on the difficult $\nicefrac{1}{16}$ split with ResNet-50 and ResNet-101 respectively.
We attribute this improvement to the additional regularization provided by the boundary multi-task learning objective, as strong regularization techniques are often particularly effective in low-supervision scenarios.
The qualitative results presented in \cref{fig:results_qualitative} further illustrate that SAMTH+BoundMatch produces more precise object boundaries and improves segmentation accuracy for challenging classes such as ``wall'', ``pole'', ``train'', and ``bus''.
Our proposed simple baseline, SAMTH, also demonstrates strong performance, achieving competitive results against previous SOTA methods.
Overall, these findings establish SAMTH as a strong baseline for urban scene SS-SS, while SAMTH+BoundMatch demonstrates competitive performance against state-of-the-art methods through its effective integration of boundary awareness.

As stated previously, BoundMatch can be integrated with other SS-SS methods, and we have shown results for UniMatch \cite{Yang2022UniMatch} and PrevMatch \cite{Shin2024PrevMatch} in \cref{tab:benchmark_cityscapes}.
For a fair comparison, we reproduced UniMatch and PrevMatch under the same experimental settings as our SAMTH baseline.
Incorporating BoundMatch into these methods, we observe consistent performance gains across all splits and backbones.
However, UniMatch and PrevMatch already leverage strong consistency regularization techniques which incurs multiple loss functions and may reduce the benefits of the boundary multi-task learning.
In future works, we will explore approaches to better balance the regularization signals, but we believe that the results shown here proves the effectiveness of BoundMatch in enhancing SS-SS methods.

Additionally, we reproduced CFCG \cite{Li2023CFCG} under our experimental settings to ensure a fair comparison, as the original benchmark partitions may differ from ours.
Our results show that SAMTH+BoundMatch outperforms CFCG and the recent CW-BASS \cite{Tarubinga2025CWBASS}, highlighting the advantages of our approach.
While CFCG and CW-BASS identify boundary regions as difficult areas and adjust loss weighting accordingly (using Laplacian and Sobel operators respectively), they do not aim to improve boundary quality directly.
Our method takes a fundamentally different approach: we explicitly learn semantic boundaries through a dedicated auxiliary task and use these learned boundaries not just for identifying difficult regions, but for actively improving both segmentation and boundary predictions through bidirectional fusion modules.
This creates a virtuous cycle where better boundaries lead to better segmentation and vice versa.

\subsubsection{BDD100K, SYNTHIA, and ADE20K}
\label{sec:benchmark_etc}

To assess the generalizability of our approach, we extended the evaluation to additional datasets: the urban driving scene datasets BDD100K and SYNTHIA, as well as the challenging academic benchmark ADE20K, known for its diversity and difficulty in semantic segmentation.
As shown in \cref{tab:benchmark_etc,tab:boundary_other}, SAMTH+BoundMatch consistently improves over supervised baselines and UniMatch on these datasets.
For instance, on BDD100K, our method improves the mean IoU by over 3\% for the $\nicefrac{1}{64}$ split.
These results demonstrate that our boundary-aware learning framework leverages unlabeled data across various domains, including challenging datasets with complex scenes.
For qualitative visualizations, please refer to \cref{fig:app_bdd100k_qualitative} in the supplementary material, which illustrates the improved segmentation boundaries and overall performance of SAMTH+BoundMatch compared to UniMatch.

\subsubsection{Pascal VOC 2012}
\label{sec:benchmark_voc2012}

We further evaluate our method on the Pascal VOC 2012 dataset using both the \textit{Classic} and \textit{Blender} splits, acknowledging the challenges posed by this benchmark.
It is important to note that Pascal VOC 2012 boundary annotations are known to be noisy, and boundary regions are typically treated as ``ignore labels'' during training and evaluation (see Appendix~\ref{app:voc_qualitative}).
Therefore, in this benchmark, we hypothesized that the gains from BoundMatch will be limited compared to the previous high-quality datasets.
However, experiment results demonstrate that BoundMatch can still provide improvements in segmentation performance, especially in the low labeled data regime as shown in this section.
We believe this is due to the strong regularization of using multi-task CR objectives with label-refinement strategy potentially improving noise robustness as theorized in \cite{Zhang2024NRCR}.

\noindent\textbf{Classic Split:} \cref{tab:benchmark_voc_classic,tab:boundary_voc_classic} presents the performance comparison on the Classic splits for both ResNet-50 and ResNet-101 backbones.
While SAMTH+BoundMatch improves performance relative to earlier CR methods like AugSeg and UniMatch, it does not reach the state-of-the-art results achieved by the most recent approaches, such as PrevMatch and CW-BASS, on this dataset.
We hypothesize this performance gap is partly attributable to the weak CR approach of SAMTH, making it less effective for this dataset.
To further assess BoundMatch's boundary mechanism independently of our SAMTH baseline, we integrated it with UniMatch and PrevMatch again.
When combined with these strong baselines, BoundMatch consistently yields mean IoU gains over the original methods, achieving results that are competitive with, or reach, current state-of-the-art performance (\cref{tab:benchmark_voc_classic}).

\noindent\textbf{Blended Split:} This split contains potentially even more noisy GT labels, particularly around object boundaries.
\cref{tab:benchmark_voc_blender,tab:boundary_voc_blender} shows trends similar to the Classic split.
Again, the combinations of BoundMatch with UniMatch or PrevMatch achieve performance competitive with recent SOTA methods.
Furthermore, BoundMatch outperforms CFCG \cite{Li2023CFCG} in this split, demonstrating its robustness even in the presence of noisy boundary annotations.

Collectively, these Pascal VOC results underscore the benefit of incorporating explicit boundary modeling in SS-SS frameworks.
While BoundMatch does not consistently outperform the best methods on this dataset, it demonstrates its potential to enhance segmentation performance when integrated with strong SS-SS baselines attributed to its simple yet effective boundary-aware learning mechanism.

\subsubsection{Recent Benchmarks with Transformer Backbones}
\label{sec:benchmark_transformer}

While DeepLabV3+ with ResNet backbones has been the standard benchmark for SS-SS methods, recent works have begun exploring vision transformers to leverage their superior representation learning capabilities \cite{yang2025unimatchv2}.
Following this direction, we evaluate BoundMatch using DPT \cite{Ranftl2021DPT} with DINOv2 \cite{Oquab2023DINOv2} pretrained weights, a foundation model that has demonstrated strong performance across various vision tasks.
Adapting BoundMatch to DPT requires minimal modifications, with implementation details provided in Appendix~\ref{app:boundmatch_dpt}.

As shown in \cref{tab:benchmark_dinov2}, SAMTH+BoundMatch outperforms current state-of-the-art methods, UniMatch-V2 \cite{yang2025unimatchv2} and SegKC \cite{Than2025SegKC}, across both DINOv2-S and DINOv2-B backbones on the challenging $\nicefrac{1}{16}$ and $\nicefrac{1}{8}$ splits.
These results demonstrate that BoundMatch scales to modern foundation models while maintaining its boundary-aware advantages.
Qualitative visualizations are provided in Appendix~\ref{app:dino_qualitative}.

\begin{table*}[t]
\centering
\caption{Component analysis of our framework.}
\label{tab:ablation_components}
\footnotesize
\begin{tabular}{
  c c c
  S[table-format=2.1] S[table-format=2.1] S[table-format=2.1] 
  S[table-format=2.1] S[table-format=2.1] S[table-format=2.1] 
  S[table-format=2.1] S[table-format=2.1] S[table-format=2.1] 
  S[table-format=2.1] S[table-format=2.1] S[table-format=2.1] 
}
\toprule
\multicolumn{3}{c}{} &
\multicolumn{6}{c}{\textbf{ResNet-50}} &
\multicolumn{6}{c}{\textbf{ResNet-101}} \\
\cmidrule(lr){4-9}\cmidrule(l){10-15}
\multicolumn{3}{c}{} &
\multicolumn{3}{c}{$\nicefrac{1}{16}$} &
\multicolumn{3}{c}{$\nicefrac{1}{8}$} &
\multicolumn{3}{c}{$\nicefrac{1}{16}$} &
\multicolumn{3}{c}{$\nicefrac{1}{8}$} \\
\cmidrule(lr){4-6}\cmidrule(lr){7-9}\cmidrule(lr){10-12}\cmidrule(l){13-15}
\textbf{BCRM} & \textbf{BSF} & \textbf{SGF} &
\multicolumn{1}{c}{IoU} & \multicolumn{1}{c}{BIoU} & \multicolumn{1}{c}{BF1} &
\multicolumn{1}{c}{IoU} & \multicolumn{1}{c}{BIoU} & \multicolumn{1}{c}{BF1} &
\multicolumn{1}{c}{IoU} & \multicolumn{1}{c}{BIoU} & \multicolumn{1}{c}{BF1} &
\multicolumn{1}{c}{IoU} & \multicolumn{1}{c}{BIoU} & \multicolumn{1}{c}{BF1} \\
\midrule
--     & --     & --     &
75.1 & 54.8 & 59.3 &
77.3 & 56.8 & 61.8 &
75.5 & 55.8 & 61.4 &
77.9 & 57.9 & 63.1 \\
\cmark & --     & --     &
75.9 & 55.3 & 60.2 &
77.6 & 56.9 & 62.3 &
76.9 & 56.9 & 62.1 &
78.2 & 58.5 & 64.5 \\
\cmark & \cmark & --     &
76.2 & 56.0 & 62.0 &
77.5 & 57.2 & 62.8 &
76.1 & 57.3 & 64.5 &
78.2 & 59.0 & 65.0 \\
\cmark & --     & \cmark &
76.1 & 55.7 & 61.6 &
77.6 & 57.0 & 62.4 &
76.6 & 57.0 & 63.9 &
77.9 & 58.8 & 64.9 \\
\rowcolor{ablrow}
\cmark & \cmark & \cmark &
\textbf{76.5} & \textbf{56.1} & \textbf{62.7} &
\textbf{78.1} & \textbf{58.1} & \textbf{64.7} &
\textbf{77.9} & \textbf{58.3} & \textbf{65.2} &
\textbf{79.0} & \textbf{59.4} & \textbf{66.7} \\
\bottomrule
\end{tabular}
\vspace{-10pt} 
\end{table*}

\subsection{Ablation Studies and Insights}
\label{sec:ablation}

\subsubsection{Individual efficacy of the proposed components}
\label{sec:component_analysis}

To quantify the individual contributions of each proposed component, we conduct systematic ablation experiments evaluating the Boundary Consistency Regularization Module (BCRM), Boundary-Semantic Fusion (BSF), and Spatial Gradient Fusion (SGF) modules.
\cref{tab:ablation_components} reports mIoU, BIoU, and BF1 across multiple evaluation protocols.

The addition of our core BCRM framework alone results in notable gains in both segmentation (0.3--1.4\%) and boundary quality metrics (0.1--1.1\% for BIoU and 0.5--1.4\% for BF1).
Further incorporating the BSF module builds upon this, enhancing performance by integrating learned boundary cues into the segmentation head.
BCRM with SGF has similar gains, indicating that refining boundary predictions using segmentation features is also beneficial.
Finally, using both BSF and SGF modules provide complementary benefits; its strategy of refining boundary predictions using spatial gradients derived from the segmentation mask leads to the highest overall performance across all protocols (0.8--2.4\% gains for mIoU).

\subsubsection{Analysis of Hyperparameters}
\label{sec:hyperparameter_analysis}

\begin{figure}[t]
  \centering
  \subfloat[$\tau_{bdry}$\label{fig:tau_bdry}]%
  {\includegraphics[width=.45\linewidth]{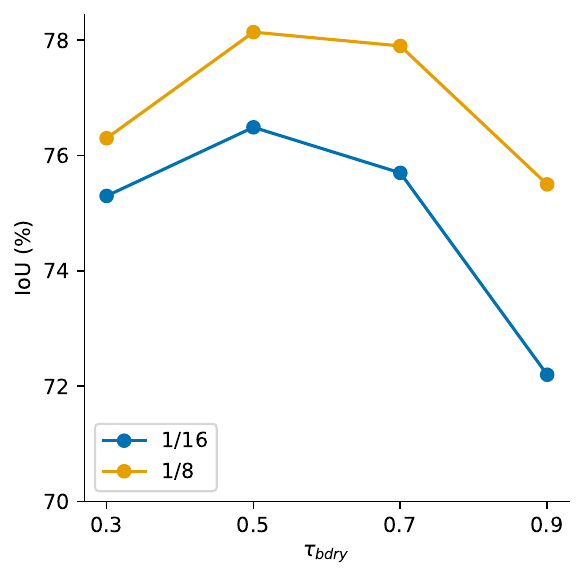}}
  \hfill
  \subfloat[$\lambda_{bdry}$\label{fig:lambda_bdry}]%
  {\includegraphics[width=.45\linewidth]{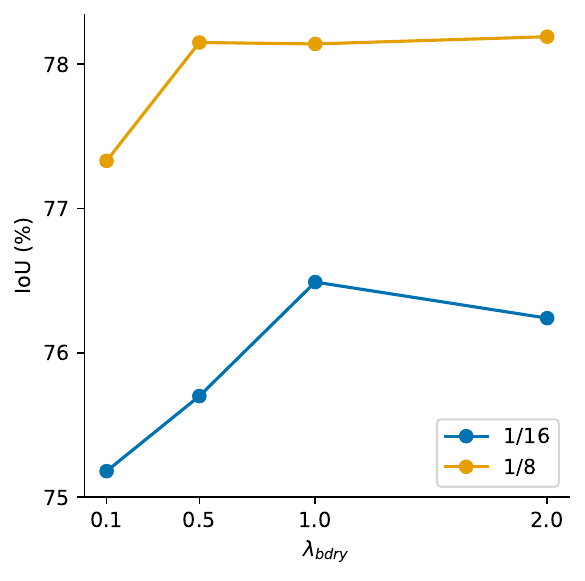}}
  \caption{Analysis of the hyperparameters introduced by BoundMatch.
           (a) Varying the boundary threshold $\tau_{bdry}$; 
           (b) Varying the boundary-loss weight $\lambda_{bdry}$. 
           For (b) we set $\tau_{bdry}=0.5$. 
           Results are on Cityscapes with the $\nicefrac{1}{16}$ 
           and $\nicefrac{1}{8}$ splits using the ResNet-50 backbone.}
  \label{fig:hyperparameter_analysis}
\vspace{-10pt} 
\end{figure}

\begin{figure*}[t]
  \centering
  \subfloat[Per-class IoU results\label{fig:per_class_iou}]%
  {\includegraphics[width=0.95\linewidth]{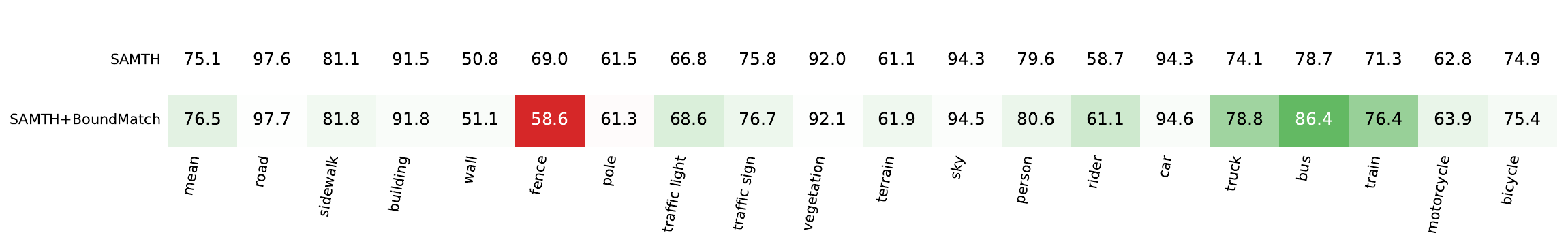}}
  \vspace{-1.0em} 
  \subfloat[Per-class BIoU results\label{fig:per_class_biou}]%
  {\includegraphics[width=0.95\linewidth]{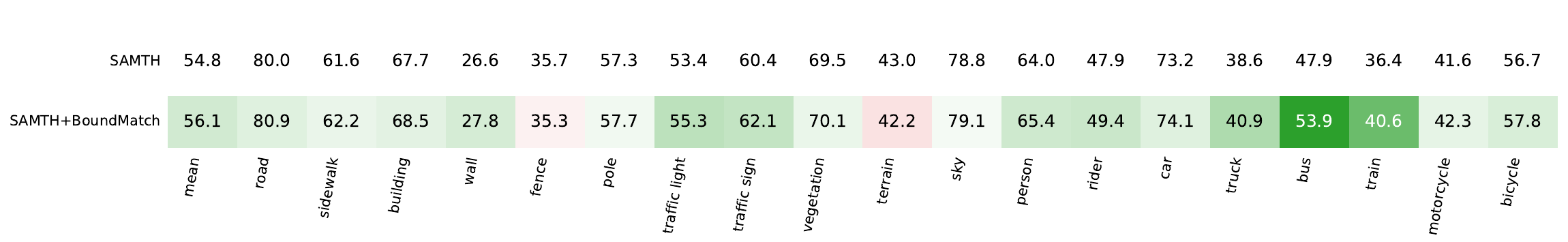}}
  \vspace{-1.0em}
  \subfloat[Per-class BF1 results\label{fig:per_class_bf1}]%
  {\includegraphics[width=0.95\linewidth]{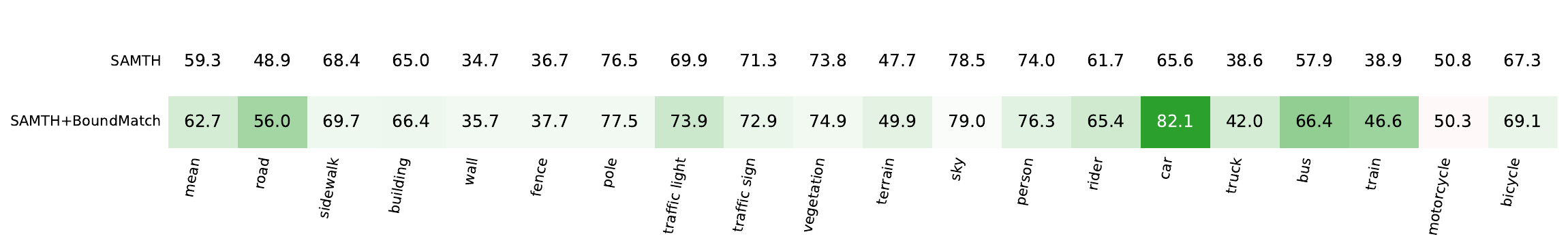}}
  \caption{Per-class performance comparison between SAMTH and
    SAMTH\,+\,BoundMatch on the Cityscapes dataset
    (\nicefrac{1}{16} split, ResNet-50 backbone).
    \textcolor{green}{Green} values denote improvements,
    \textcolor{red}{red} values denote degradation.}
  \label{fig:per_class_performance}
\end{figure*}

\begin{figure*}[!t]
  \centering
  \begin{minipage}{0.41\textwidth}
    \centering
    \includegraphics[width=\linewidth]{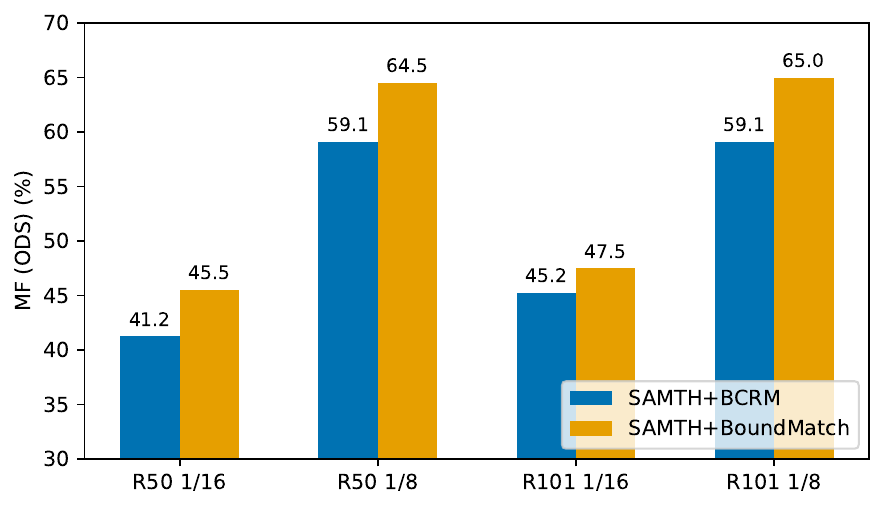}
  \end{minipage}
  \hfill
  \begin{minipage}{0.58\textwidth}
    \centering
    \includegraphics[width=\linewidth]{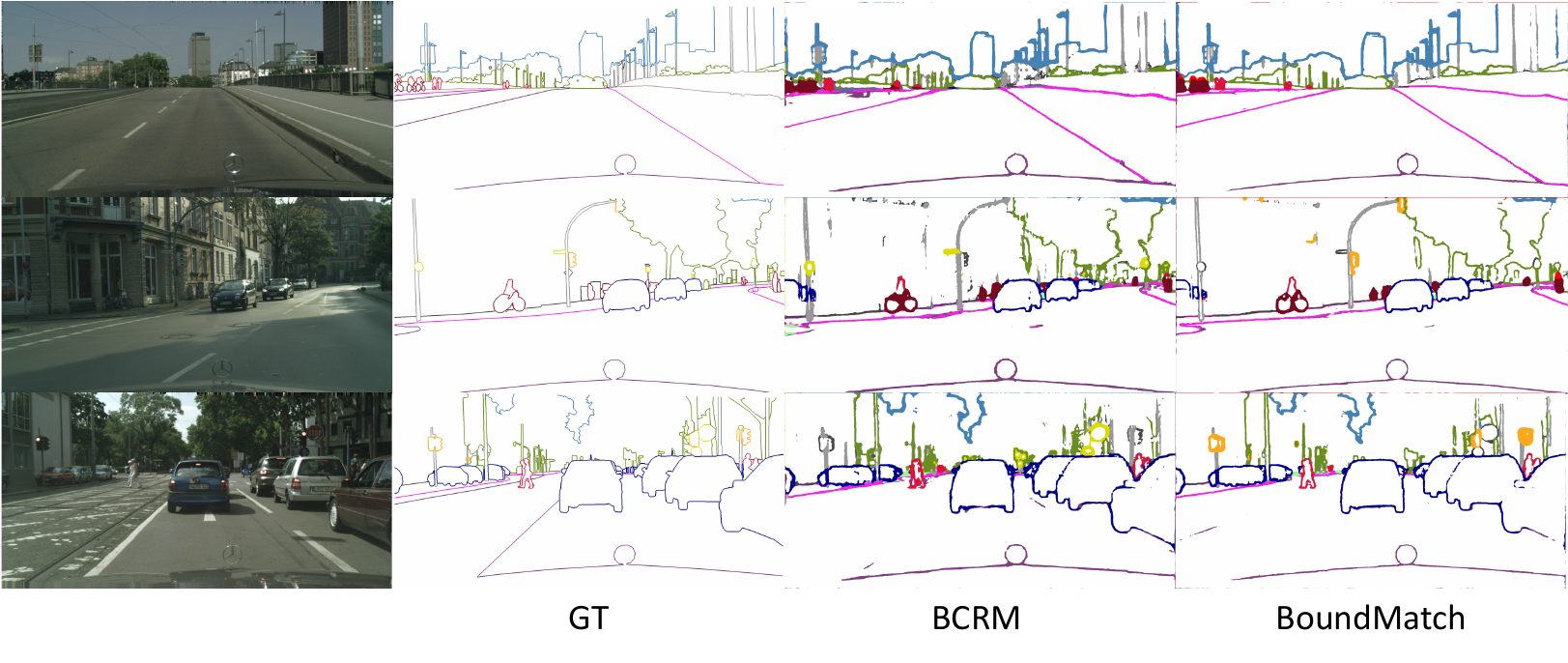}
  \end{minipage}
  \caption{Comparison of semantic boundary predictions. (a) MF (ODS) score of boundary predictions for BCRM versus BoundMatch. (b) Qualitative comparison illustrating the impact of Spatial Gradient Fusion (SGF) on boundary predictions (Cityscapes dataset, $\nicefrac{1}{16}$ split). Boundaries generated using only the BCRM component tend to be thicker and contain more artifacts (false positives). In contrast, the full BoundMatch framework, incorporating the SGF module, produces visibly sharper and cleaner boundaries. SGF refines the initial boundary predictions by leveraging spatial gradient information derived from the segmentation mask, yielding higher-quality boundary pseudo-labels that enhance the effectiveness of consistency regularization.}
  \label{fig:ablation_boundaries}
  \vspace{-5pt}
\end{figure*}

Having established the efficacy of individual components, we now examine the sensitivity to key hyperparameters.
\cref{fig:hyperparameter_analysis} presents the impact of two key hyperparameters introduced by BoundMatch: the boundary threshold $\tau_{bdry}$ and boundary loss weight $\lambda_{bdry}$ on the Cityscapes dataset using ResNet-50 backbone.
For the boundary threshold, we find that $\tau_{bdry} = 0.5$ yields optimal performance across both data splits.
This moderate threshold balances two factors: boundary predictions typically have low probability values due to the sparsity of boundary pixels, yet setting the threshold too high would exclude valuable boundary information from the consistency regularization process.

Regarding the boundary loss weight $\lambda_{bdry}$, optimal performance is achieved with values around $\lambda_{bdry}=1.0$ for both splits.
Interestingly, the $\nicefrac{1}{8}$ split shows greater robustness to variations in $\lambda_{bdry}$ compared to the $\nicefrac{1}{16}$ split, suggesting that increased labeled data provides more stable training dynamics that are less sensitive to the precise weighting of the boundary loss.

\subsubsection{Binary and Semantic Boundaries}
\label{sec:binary_vs_semantic}

\begin{table}[t]
\centering
\caption{Binary vs. Multi-Label Boundaries. "Derived" means the model uses boundaries are derived from the predicted pseudo-labels from the segmentation head. "Learned" means predicted boundaries are directly used as the supervision signal.}
\label{tab:ablation_bdry_types}
\footnotesize
\setlength{\tabcolsep}{4pt}

\begin{tabular}{
  l
  S[table-format=2.1] S[table-format=2.1] S[table-format=2.1] 
  S[table-format=2.1] S[table-format=2.1] S[table-format=2.1] 
}
\toprule
& \multicolumn{3}{c}{$\nicefrac{1}{16}$} & \multicolumn{3}{c}{$\nicefrac{1}{8}$}\\
\cmidrule(lr){2-4}\cmidrule(lr){5-7}
& {IoU} & {BIoU} & {BF1} & {IoU} & {BIoU} & {BF1}\\
\midrule
SAMTH        & 75.1 & 54.8 & 59.3 & 77.3 & 56.8 & 61.8\\
\quad + Binary (Derived) & 75.3 & 54.6 & 59.4 & 77.3 & 56.9 & 62.0\\
\rowcolor{ablrow}
\quad + Binary (Learned)     & 75.9 & 55.4 & 60.1 & 77.6 & 57.3 & 62.5\\ 
\quad + Multi-Label (Derived) & 75.8 & 55.4 & 60.8 & 77.4 & 57.1 & 62.5 \\
\rowcolor{ablrow}
  \quad + Multi-Label (Learned) & \textbf{76.5} & \textbf{56.1} & \textbf{62.7} & \textbf{78.1} & \textbf{58.1} & \textbf{64.7} \\
\bottomrule
\end{tabular}
\vspace{-10pt} 
\end{table}

An important design decision in our method is what types of boundaries should be used: derived versus learned, and binary versus semantic (multi-label).
As shown in \cref{tab:ablation_bdry_types}, using learned multi-label boundaries obtains the highest improvements across both evaluation protocols, outperforming baseline with binary boundaries and derived boundaries from segmentation pseudo-labels.
For $\nicefrac{1}{16}$, IoU improves by $1.4\%$, BIoU by $1.3\%$, and BF1 by $3.4\%$, indicating that learned class-aware boundary supervision provides benefits over the baseline approach.

Using derived boundaries show minimal improvements over SAMTH compared to using learned boundaries for consistency regularization.
We believe this is because generating boundaries (\eg via Laplacian operators) from often noisy segmentation pseudo-labels propagates these noise to the boundary predictions, limiting their effectiveness.
In contrast, learning boundaries directly from dedicated boundary predictions allows the model to focus on the geometric structures of the boundaries independent of the segmentation task, and the ability to threshold boundaries by confidence ($\tau_{bdry}$) allows the model to learn more robust boundary representations.
Furthermore, our bidirectional fusion allows the boundary and segmentation heads to mutually benefit from each other, where SGF module enables learning with segmentation mask cues.

\subsubsection{Per-Class Performance}
\label{app:per_class_performance}

To understand how these boundary formulations affect different object categories, we analyze per-class performance.
Here we present the per-class performance comparison between SAMTH and BoundMatch for Cityscapes $\nicefrac{1}{16}$ split in \cref{fig:per_class_iou,fig:per_class_biou,fig:per_class_bf1} for IoU, BIoU, and BF1 metrics respectively.
For IoU, we observe that BoundMatch improves the performance on most of the classes, especially on vehicles (\eg car, truck, bus, and train) and people (\ie person and rider).
While thin and small objects (\eg traffic light and traffic sign) also see improvements, SAMTH+BoundMatch seems to have less impact on pole although we see qualitative improvements for poles in \cref{fig:app_successful_cases}
This may be due to pixel imbalance issues, as poles occupy a very small fraction of the image area, but it could also be caused by noisy annotations in the ground-truth which amplifies false positives as shown in \cref{fig:app_failure_cases} due to BoundMatch focusing more on the details.
The fence category presents an interesting exception with decreased IoU but maintained boundary metrics, likely due to annotation ambiguities where fences are labeled as solid regions despite their sparse structure as shown in \cref{fig:fence_results}.
BoundMatch improves BIoU and BF1 metrics for most object categories, though minor degradations are observed for certain classes with ambiguous boundaries (\eg fence, terrain), which aligns with the challenges discussed in our limitations.

\subsubsection{Boundary Evaluation}
\label{sec:boundary_evaluation}

\cref{fig:ablation_boundaries} presents quantitative and qualitative results comparing boundary predictions from BCRM alone versus the full BoundMatch framework (BCRM+BSF+SGF).
Looking at the mean F-measure (MF) at optimal dataset scale (ODS), we observe that BoundMatch consistently outperforms BCRM alone across various experiment protocols, indicating that the SGF refinement module improves boundary quality.
Qualitatively, BCRM alone produces thicker boundaries with more artifacts, while the full BoundMatch framework generates sharper and cleaner boundaries.
This improvement demonstrates the benefit of SGF's spatial gradient fusion in refining boundary predictions for more reliable pseudo-labels.

\subsubsection{Comparisons with BoundaryMatch}
\label{sec:comparison_boundarymatch}

\begin{table}[t]
\centering
\caption{Comparison between our BoundMatch and BoundaryMatch \cite{Li2024BoundaryMatch}. BoundaryMatch$^\dag$ is our reproduced results. All models use DeepLabV3$+$ with ResNet-101.}
\label{tab:ablation_boundarymatch}
\footnotesize
\setlength{\tabcolsep}{3.2pt}

\begin{tabular}{
  l
  S                      
  S S S                  
}
\toprule
\multirow{2}{*}{\textbf{Methods}} &
\multicolumn{1}{c}{\textbf{VOC (92)}} &
\multicolumn{3}{c}{\textbf{Cityscapes ($\nicefrac{1}{16}$)}} \\
\cmidrule(lr){3-5}
& {IoU} & {IoU} & {BIoU} & {BF1} \\
\midrule
BoundaryMatch          & 75.4 & 76.0 & {--} & {--} \\
\midrule
UniMatch                       & 75.2 & 76.6 & 57.0 & 63.4 \\
BoundaryMatch$^\dag$        & 75.5 & 76.5 & 56.7 & 63.5 \\
\quad + Multi-Label & 75.5 & 76.7 & 57.1 & 63.6\\
\rowcolor{ablrow}
UniMatch + BoundMatch (ours) & 76.0 & 77.4 & 58.0 & 64.5 \\
\rowcolor{ablrow}
SAMTH + BoundMatch (ours)   & \textbf{76.6} & \textbf{77.9} & \textbf{58.5} & \textbf{65.3} \\
\bottomrule
\end{tabular}
\vspace{-10pt} 
\end{table}

We compared BoundMatch against BoundaryMatch \cite{Li2024BoundaryMatch}, a recent method that also utilizes boundary CR for SS-SS.
For a fair comparison based on our experimental setup, we reimplemented BoundaryMatch and trained it using identical settings.
Quantitative results on Pascal VOC and Cityscapes dataset using ResNet-101 backbone are presented in \cref{tab:ablation_boundarymatch} which also includes the reported mIoU for BoundaryMatch.

BoundaryMatch achieves higher mIoU compared to UniMatch baseline in Pascal VOC, but does not seem to improve IoU on Cityscapes.
With multi-label boundaries, the metrics does surpass UniMatch slightly.
UniMatch+BoundMatch achieves 0.9\% mIoU and SAMTH+BoundMatch achieves an even higher 1.4\% compared to BoundaryMatch on Cityscapes.
Furthermore, boundary metrics (BIoU and BF1) also show notable improvements with BoundMatch, indicating that our method is more effective in capturing boundary details, while BoundaryMatch has similar boundary metrics to UniMatch.

BoundMatch's improved performance stems from four key technical advantages that address the limitations of simpler consistency approaches:
(1) \emph{semantic} boundary supervision, providing richer class-specific edge cues compared to binary boundaries;
(2) the use of \emph{learned} boundaries as pseudo-labels for consistency regularization;
(3) boundary heads leveraging \emph{hierarchical features} from the backbone with \emph{deep supervision}, known to be effective for capturing multi-scale boundary details \cite{yu2017casenet}; and (4) dedicated \emph{fusion modules} (BSF, SGF) that integrate boundary information back into the segmentation process and refine boundary predictions.
This suggests that simpler boundary consistency regularization might not fully exploit the potential of boundary cues, whereas our multi-faceted approach proves more effective in the SS-SS context.

\subsubsection{Effect of Harmonious BN update strategy}
\label{sec:effect_HBN}

\begin{table}[t]
\centering
\caption{Effect of Harmonious Batch-Norm (HBN) update on Pascal VOC (92 labeled image) and Cityscapes ($\nicefrac{1}{16}$ split) datasets using ResNet-50 backbone. We report the mean IoU and the absolute gains ($\Delta$) after applying HBN to various base methods.}
\label{tab:ablation_hbn}
\footnotesize
\setlength{\tabcolsep}{4pt}

\begin{tabular}{
  l
  S  S[table-format = +2.1]   
  S  S[table-format = +2.1]   
}
\toprule
\multirow{2}{*}{\textbf{Methods}} &
\multicolumn{2}{c}{\textbf{VOC (92)}} &
\multicolumn{2}{c}{\textbf{Cityscapes ($\nicefrac{1}{16}$)}} \\
\cmidrule(lr){2-3}\cmidrule(l){4-5}
& {IoU} & {$\Delta$} & {IoU} & {$\Delta$} \\
\midrule
Mean-Teacher                & 51.7 & {}        & 66.1 & {}        \\
\rowcolor{ablrow}
\quad + HBN                 & 61.5 & {+9.8}      & 71.7 & {+5.6}      \\

CutMix-MT                   & 52.2 & {}        & 68.3 & {}        \\
\rowcolor{ablrow}
\quad + HBN                 & 68.6 & {+16.4}     & 73.2 & {+4.9}     \\

ReCo                        & 64.8 & {}        & 71.4 & {}        \\
\rowcolor{ablrow}
\quad + HBN                 & 67.5 & {+2.7}      & 75.1 & {+3.7}      \\

U2PL                        & 68.0 & {}        & 70.3 & {}        \\
\rowcolor{ablrow}
\quad + HBN                 & 69.1 & {+1.1}      & 72.4 & {+2.1}      \\

iMAS                        & 68.8 & {}        & 74.3 & {}        \\
\rowcolor{ablrow}
\quad + HBN                 & 69.7 & {+0.9}      & 75.1 & {+0.8}      \\

AugSeg                      & 71.1 & {}        & 73.7 & {}        \\
\rowcolor{ablrow}
\quad + HBN                 & 74.1 & {+3.0}      & 75.1 & {+1.4}      \\

SAMTH (no HBN)              & 62.8 & {}        & 69.4 & {}        \\
\rowcolor{ablrow}
SAMTH                       & 70.7 & {+7.9}      & 75.1 & {+5.7}      \\
\bottomrule
\end{tabular}
\vspace{-10pt} 
\end{table}

\begin{figure}[t]
  \centering
   \includegraphics[width=0.8\linewidth]{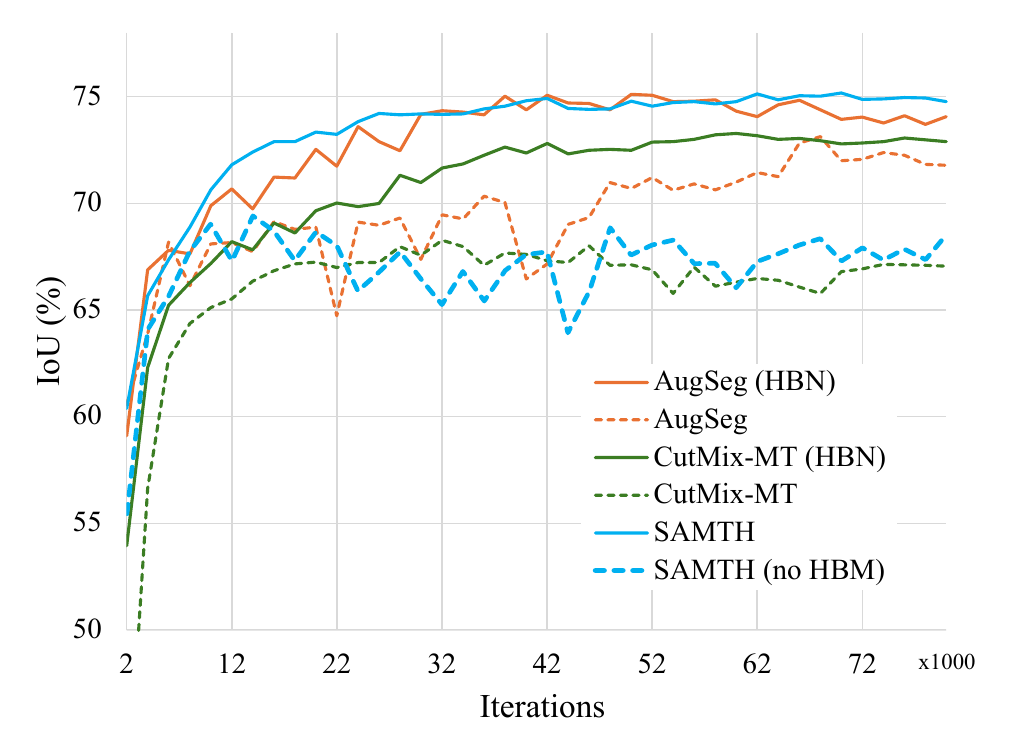}
   \caption{Training curves of AugSeg, CutMix-MT, and SAMTH on the Cityscapes $\nicefrac{1}{16}$ benchmark using ResNet-50 backbone. The bold lines are method trained with HBN update strategy, while the dotted lines are the ones without.}
   \label{fig:ablation_hbn}
\vspace{-5pt} 
\end{figure}

To further examine the effect of our Harmonious Batch Normalization (HBN) update strategy, we compare different BN update approaches in \cref{tab:ablation_hbn}.
Methods like CutMix-MT, ReCo, and U2PL also update the teacher's BN statistics via forward passes.
However, unlike HBN (which processes the complete batch through the teacher specifically for BN updates, as described in \cref{sec:samth}), these methods often introduce discrepancies between the inputs used for the teacher's BN statistics and the student's training inputs.
Other approaches, such as iMAS and AugSeg, update the teacher's BN buffers via EMA based on the student's statistics.
In our experiments, incorporating HBN improves segmentation performance for most tested baseline methods, with mIoU gains ranging from 0.8--16.4\%.
For example, applying HBN to SAMTH results in a 5.7\% absolute improvement in mean IoU on Cityscapes (from 69.4\% to 75.1\%).
The training curves in \cref{fig:ablation_hbn} provide further evidence, indicating that not only does HBN results in higher final performance, but it also stabilizes the training process.

HBN is not the main focus of this work, but it is a simple yet effective strategy that can be applied to any SS-SS method that uses a teacher-student framework with EMA updates.
It led SAMTH, a simple SDA method, to achieve competitive performance on the benchmarks.

\subsubsection{Computational Cost}
\label{sec:computational_cost}

\begin{table}[t]
\centering
\caption{Computation cost comparison. We report the average iteration time in seconds and the GPU memory required by a single GPU during \textbf{training}. We also report FLOPs, number of parameters, and FPS during \textbf{inference} using NVIDIA RTX 3090 GPU.}
\label{tab:ablation_computational_cost}
\scriptsize
\setlength{\tabcolsep}{3pt}

\begin{tabular}{
  l               
  S S S           
  S[table-format=1.2, table-space-text-post=\,s]  
  S[table-format=2.1, table-space-text-post=\,GB]  
  S[table-format=3.0, table-space-text-post=\,G]  
  S[table-format=2.1, table-space-text-post=\,M]  
  S                 
}
\toprule
\multirow{2}{*}{\textbf{Method}} &
\multicolumn{3}{c}{\textbf{Accuracy (\%)}} &
\multicolumn{2}{c}{\textbf{Training cost}} &
\multicolumn{3}{c}{\textbf{Inference cost}} \\
\cmidrule(lr){2-4}\cmidrule(lr){5-6}\cmidrule(l){7-9}
& {IoU} & {BIoU} & {BF1} & {Time} & {Mem} & {FLOPs} & {Params} & {FPS} \\
\midrule
UniMatch                     & 75.3 & 55.1 & 61.0 & 1.21s & 20.6GB & 191G & 40.5M & 23.4\\
PrevMatch                    & 75.7 & 55.3 & 60.7 & 1.56s & 21.4GB & 191G & 40.5M & 23.4\\

\rowcolor{ablrow}
SAMTH                        & 75.1 & 54.8 & 59.3 & 0.92s &  9.0GB & 191G & 40.5M & 23.4\\
\rowcolor{ablrow}
+ BCRM                       & 75.9 & 55.3 & 60.2 & 1.24s & 13.2GB & 191G & 40.5M & 23.4\\
\rowcolor{ablrow}
+ BCRM + BSF                 & 76.2 & 56.0 & 62.0 & 1.32s & 13.5GB & 192G & 40.6M & 19.0\\
\rowcolor{ablrow}
+ BCRM + SGF                 & 76.1 & 55.7 & 61.6 & 1.45s & 15.0GB & 191G & 40.5M & 23.4\\
\rowcolor{ablrow}
+ BoundMatch           & 76.5 & 56.1 & 62.7 & 1.59s & 16.5GB & 192G & 40.6M & 19.0 \\
\bottomrule
\end{tabular}
\vspace{-10pt} 
\end{table}

\cref{tab:ablation_computational_cost} analyzes the computational overhead of BoundMatch's individual components.
BCRM adds 0.32s and 4.2GB for multi-scale boundary processing, BSF contributes minimal overhead (0.08s, 0.3GB), while SGF requires 0.21s and 1.8GB for gradient computations.
The complete framework efficiently combines these modules with shared computations, resulting in a total training overhead of 0.67s and 7.5GB compared to SAMTH baseline, primarily due to the on-the-fly boundary generation pipeline \cite{ishikawa2023SBCB} and the fusion modules.
Despite this increase, BoundMatch remains more memory-efficient than UniMatch (20.6GB) and have similar per-iteration time as PrevMatch (1.56s) while achieving accuracy gains.
Optionally, if training costs are a concern, one can disable BSF and only use BCRM+SGF which still provides 1.0\% mIoU improvement over SAMTH with only 0.53s and 6.0GB overhead.

At inference, BoundMatch introduces minimal overhead (1G FLOPs, 0.1M parameters), reducing FPS from 23.4 to 19.0 for 1.4\% mIoU and up to 3.4\% boundary metric improvements.
BSF slightly increases FLOPs and reduces FPS due to additional convolutions required feature fusion, thus if the trade-off is not desired, one can use the BCRM+SGF combination which has no impact on inference cost.
BoundMatch's modular design allows users to change feature fusion modules entirely (\eg use attention-based fusion), which may offer different trade-offs between accuracy and efficiency.
Investigations into better feature fusion strategies that are computationally efficient are left for future work.

\subsection{More Real-World SS-SS Setting}
\label{sec:real_world_setting}

\begin{table}[t]
\centering
\caption{Evaluation of methods in a more real-world semi-supervised setting where we use the entire labeled Cityscapes dataset (2975 images) along with the \textit{extra} 19997 unlabeled training data. We evaluate under the same setting for all methods and use ResNet-50 as the backbone.}
\label{tab:real_world_setting}
\footnotesize

\begin{tabular}{
  l
  S[table-format=2.2]
  S[table-format=2.2]
  S[table-format=2.2]
}
\toprule
\textbf{Method} & {IoU} & {BIoU} & {BF1} \\ \midrule
Supervised only     & 77.58 & 57.73 & 60.06 \\
UniMatch            & 80.18 & 61.18 & 67.43 \\ \midrule
SAMTH               & 80.26 & 60.86 & 65.21 \\
\rowcolor{ablrow}
\quad + BoundMatch  & 80.83 & 62.24 & 69.02 \\ \bottomrule
\end{tabular}
\vspace{-10pt} 
\end{table}

Inspired by the evaluation protocol in \cite{yang2025unimatchv2}, we assess SS‑SS in a realistic scenario: using all 2,975 Cityscapes training images as labeled data and the additional 19,997 “extra” images as truly unlabeled data.
In this setting, the fully supervised baseline is already strong, making further gains challenging yet highly relevant for practical deployment.
As shown in \cref{tab:real_world_setting}, SAMTH+BoundMatch achieves consistent improvements over both SAMTH and UniMatch, particularly in boundary‑specific metrics (BIoU and BF1), demonstrating its effectiveness at leveraging large unlabeled pools in real‑world applications.

\subsection{Application to lightweight models}
\label{sec:light_weight}

\begin{table}[t]
\centering
\caption{Lightweight segmentation architectures applied to the same setting as \cref{tab:real_world_setting}. Here ``BoundMatch'' refers to SAMTH+BoundMatch which is trained with the additional unlabeled \textit{Extra} data. We use input size of $768\times 768$ and $800\times 800$ for MobileNet-V2 and AFFormer-Tiny, respectively.}
\label{tab:light_weight}
\scriptsize
\setlength{\tabcolsep}{3pt}

\begin{tabular}{
  l l
  S S S
  S[table-format=3.1, table-space-text-post={\,G}]   
  S[table-format=2.1, table-space-text-post={\,M}]   
  S                   
}
\toprule
\multicolumn{2}{c}{\textbf{Method}} &
\multicolumn{3}{c}{\textbf{Accuracy (\%)}} &
\multicolumn{3}{c}{\textbf{Efficiency}} \\
\cmidrule(lr){3-5}\cmidrule(l){6-8}
\multicolumn{2}{c}{} & {IoU} & {BIoU} & {BF1} & {FLOPs} & {Params} & {FPS} \\
\midrule
\multirow{2}{*}{\textbf{DLV3+ MV2}}
  & Baseline     & 73.4 & 53.9 & 56.5 & 89.7G & 5.8M & 36.0 \\
  & \cellcolor{ablrow} BoundMatch & \cellcolor{ablrow} 78.2 & \cellcolor{ablrow} 59.0 & \cellcolor{ablrow} 64.3 &
     \cellcolor{ablrow} 93.7G & \cellcolor{ablrow} 6.0M & \cellcolor{ablrow} 31.8 \\

\addlinespace[2pt]
\multirow{3}{*}{\textbf{AFFormer-Tiny}}
  & Baseline     & 76.7 & 57.5 & 65.2 &  7.3G & 2.1M & 89.1 \\
  & Mobile-Seed  & 78.0 & 58.3 & 65.1 & 10.0G & 2.4M & 70.2 \\
  & \cellcolor{ablrow} BoundMatch & \cellcolor{ablrow} 80.1 & \cellcolor{ablrow} 61.1 & \cellcolor{ablrow} 68.2 &
     \cellcolor{ablrow} 9.2G & \cellcolor{ablrow} 2.2M & \cellcolor{ablrow} 78.0 \\
\bottomrule
\end{tabular}
\vspace{-10pt} 
\end{table}

We evaluate BoundMatch on lightweight architectures suitable for resource-constrained deployment, specifically DeepLabV3+ with MobileNet-V2 \cite{Sandler2018MobileNetV2} and AFFormer-Tiny \cite{Dong2023AFFormer}.
For AFFormer-Tiny, we compare against Mobile-Seed \cite{liao2024mobileseed}, a state-of-the-art boundary-aware lightweight segmentation method.
More information about the experimental setup is provided in Appendix~\ref{app:lightweight_models}.

As shown in \cref{tab:light_weight}, BoundMatch achieves accuracy improvements of 3.4--4.8\% mIoU with moderate impact on runtime performance, maintaining above 30 FPS for real-time applications.
For MobileNet-V2, we achieve 4.8\% IoU and 7.8\% boundary F1 gains with only 4.5\% additional FLOPs, maintaining 31.8 FPS suitable for real-time applications.
More notably, AFFormer-Tiny with BoundMatch reaches 80.1\% IoU and 61.1\% BIoU while sustaining 78 FPS, well above the 30 FPS threshold required for real-time video processing.
With only 2.2M parameters and 9.2 GFLOPs, this configuration is particularly suited for mobile deployment where memory and power constraints are important considerations.

Compared to the specialized Mobile-Seed method, BoundMatch achieves superior accuracy with 8\% fewer FLOPs and 11\% higher throughput, demonstrating that our multi-task learning framework with semi-supervised training provides a more efficient solution than dedicated boundary architectures.
The modest computational overhead (approximately 26\% additional FLOPs for AFFormer-Tiny) is justified by boundary quality improvements important for applications such as autonomous navigation, medical imaging on portable devices, and augmented reality systems.

These results confirm that BoundMatch scales to lightweight architectures without compromising deployment feasibility, making it practical for enhancing segmentation quality in resource-constrained environments where both accuracy and efficiency are important.

\subsection{Limitations and Future Work}
\label{sec:limitation}

\begin{figure}[t]
  \centering
  \includegraphics[width=\linewidth]{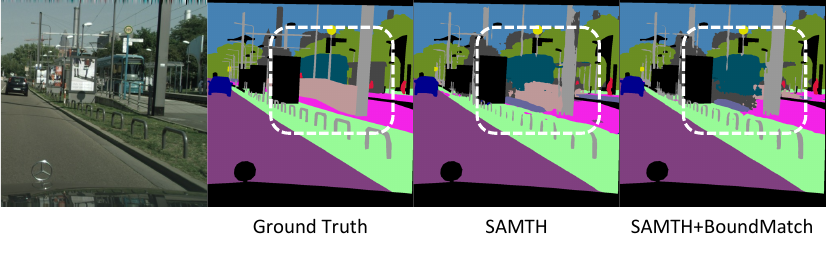}
  \vspace{-15pt} 
  \caption{An example of BoundMatch failing to segment the fence class (the tan region inside the white box). Instead, BoundMatch tries to predict the train which can be seen through the fence.}
  \label{fig:fence_results}
\vspace{-10pt} 
\end{figure}

While BoundMatch demonstrates consistent improvements across benchmarks, several limitations merit discussion and guide future research:

\noindent\textbf{Challenging structures:} Classes with fine, partially transparent, or heavily occluded structures (\eg fences, see-through facades) can degrade performance, likely due to annotation policies (filled silhouettes) and the model's tendency to favor occluded content behind meshes (\cref{fig:fence_results} and \cref{fig:app_failure_cases}).
Future work should explore attention-based modules for long-range context modeling and investigate alternative supervision strategies that better model transparency and occlusion.

\noindent\textbf{Computational considerations:} Training overhead increases (iteration time 0.92$\rightarrow$1.59s, memory 9.0$\rightarrow$16.5GB) primarily due to hierarchical boundary processing and on-the-fly boundary generation.
While our modular design allows efficiency-focused configurations (\eg BCRM+SGF maintains baseline FPS), developing more efficient boundary modules through techniques like knowledge distillation or pruning could further reduce this overhead without sacrificing performance gains.

\noindent\textbf{Domain generalization:} BoundMatch shows strong performance across most evaluated domains including remote-sensing (\cref{tab:benchmark_loveda}), but does not show gains on medical imaging (\cref{tab:benchmark_acdc}).
The boundary detection components developed in this work have not been optimized for medical imaging settings, which may require different design considerations.
Future work could investigate how to adapt boundary-aware learning to diverse application domains with their specific requirements and BoundMatch may benefit from dataset-specific optimizations.

\noindent\textbf{Annotation quality dependencies:} BoundMatch's performance is fundamentally limited by the quality of the underlying segmentation annotations.
Errors or inconsistencies in segmentation masks propagate through both the training process and boundary generation via distance transforms.
This dual dependency is particularly problematic in datasets with known annotation issues (\eg Pascal VOC's ignore regions and Cityscapes' inconsistent annotations \cref{fig:app_failure_cases}) where both segmentation and derived boundaries inherit the same biases.
Future work should explore self-supervised boundary discovery or leverage foundation models' priors to reduce dependence on precise annotations.
Additionally, developing robust training strategies that explicitly account for annotation noise in both tasks could improve real-world deployment.

\noindent\textbf{Modern architecture integration:} As vision transformers and foundation models become prevalent, adapting BoundMatch's principles presents opportunities.
Incorporating boundary-aware consistency into vision-language models could enhance their spatial understanding for tasks requiring precise localization.
The multi-task consistency principle could extend to other complementary views (depth, surface normals) in semi-supervised settings, particularly as foundation models provide increasingly rich priors for such tasks \cite{Oquab2023DINOv2}.

\section{Conclusion}
\label{sec:conclusion}

This paper presented BoundMatch, a framework that integrates semantic boundary detection into semi-supervised segmentation through consistency regularization.
Rather than deriving boundaries from segmentation outputs, we learn them from hierarchical features through independent task heads, with BSF and SGF modules enabling bidirectional information flow between tasks.

Our experiments across six datasets show consistent improvements of 0.4--2.4\% mIoU alongside gains in boundary-specific metrics (BIoU and BF1).
The modular design allows flexible deployment---using all components or selecting based on efficiency requirements.
Additionally, our Harmonious Batch Normalization (HBN) addresses training instabilities in teacher-student frameworks, improving baseline methods by up to 16.4\% mIoU.

BoundMatch integrates with existing methods (UniMatch, PrevMatch), scales to modern architectures (DINOv2 transformers), and maintains benefits on lightweight models for real-time deployment.
While improvements are incremental, their consistency validates boundary-aware learning as a useful principle for semi-supervised segmentation, particularly for applications requiring precise object delineation such as autonomous driving.

Looking forward, this multi-task consistency approach could extend to other geometric cues from foundation models, while incorporating boundary awareness into vision-language models could enhance their spatial precision.
Understanding the theoretical basis for boundary-segmentation complementarity remains an important open question.

\appendix

\section{Harmonious Batch Normalization (HBN)}
\label{app:hbn}

\begin{algorithm}[ht]
\footnotesize
\caption{One training iteration with HBN.}
\label{alg:hbn}
\SetKwInOut{Input}{Input}\SetKwInOut{Output}{Output}
\Input{Student weights $\theta^{S}$, teacher weights $\theta^{T}$, mini-batch $(x, u^{w}, u^{s})$, EMA decay $\alpha$, BN momentum $\rho$}
\Output{Updated $\theta^{S}$, $\theta^{T}$}
\begin{enumerate}[leftmargin=*,itemsep=2pt]
  \item \textbf{Teacher forward pass} (\lstinline{train} mode):
        $p^{T,L}, p^{T,w}, p^{T,s} \leftarrow f_{\theta^{T}}(x, u^{w}, u^{s})$; update teacher BN buffers with $\rho$.
  \item \textbf{Generate hard pseudo-labels}: $\hat{p}^{T} \leftarrow \operatorname*{argmax} p^{T,w}_{c}$ (apply\\ confidence $\tau$ if needed).
  \item \textbf{Student forward pass} (\lstinline{train} mode, twin-view):
        $p^{S,L}, p^{S,w}, p^{S,s} \leftarrow f_{\theta^{S}}(x, u^{w}, u^{s})$; also update student BN buffers\\ with $\rho$.
  \item \textbf{Loss computation \& student update}: compute $\mathcal L^{L}$ on $(p^{S,L}, y)$ and \\ $\mathcal L^{U}$ on $(p^{S,s}, \hat p^{T})$; update $\theta^{S}$.
  \item \textbf{EMA weight update (teacher)}: $\theta^{T} \leftarrow \alpha\,\theta^{T} + (1-\alpha)\,\theta^{S}$.
\end{enumerate}
\end{algorithm}

In teacher-student consistency training the teacher weights $\theta^{T}$ are updated as an exponential moving average (EMA) of the student weights $\theta^{S}$.
Copying the student’s batch normalization (BN) running statistics, common in prior work, makes those statistics \emph{incompatible} with $\theta^{T}$, because the teacher’s parameter trajectory and input distribution differ from those of the student.
HBN fixes this by
\begin{itemize}[leftmargin=*]
  \item \textbf{Teacher recalibration:} computing and accumulating BN statistics \emph{inside the teacher’s own forward pass} on every iteration, yielding self-consistent normalization.
  \item \textbf{Student twin-view:} forwarding the weakly augmented unlabeled images $u^{w}$ through the \emph{student} as well.
$u^{w}$ is normally not used in the student forward pass and only used to generate pseudo-labels with the teacher, but this simple step aligns the stochastic input distribution seen by both networks, further reducing BN mismatch.
\end{itemize}

A mini-batch $\mathcal B = \mathcal B^{L} \cup \mathcal B^{U,w} \cup \mathcal B^{U,s}$ contains $N$ images: labeled $x\!\in\!\mathcal B^{L}$, weakly augmented unlabeled $u^{w}\!\in\!\mathcal B^{U,w}$, and strongly augmented counterparts $u^{s}\!\in\!\mathcal B^{U,s}$.
For a BN layer $\ell$ in the \emph{teacher} network $f_{\theta^{T}}$, let the activation tensor be $h^{(\ell)}\!\in\!\mathbb R^{N\times C_\ell\times H_\ell\times W_\ell}$.

During the teacher forward pass (\lstinline{train} mode) we compute
\begin{align}
  \mu_{\mathcal B}^{(\ell)} &=
    \tfrac1N\sum_{i=1}^{N} h_{i}^{(\ell)}, &
  (\sigma_{\mathcal B}^{(\ell)})^{2} &=
    \tfrac1N\sum_{i=1}^{N}
      \bigl(h_{i}^{(\ell)}-\mu_{\mathcal B}^{(\ell)}\bigr)^{2}.
\end{align}
The teacher’s running statistics are updated with BN momentum $\rho$:
\begin{align}
  \tilde\mu^{T,(\ell)} &\leftarrow (1-\rho)\,\tilde\mu^{T,(\ell)} +
  \rho\,\mu_{\mathcal B}^{(\ell)},\\[2pt]
  (\tilde\sigma^{\,T,(\ell)})^2 &\leftarrow (1-\rho)\,(\tilde\sigma^{\,T,(\ell)})^2 +
  \rho\,(\sigma_{\mathcal B}^{(\ell)})^{2}.
\end{align}
Normalization and affine transformation then proceed as usual.
Algorithm~\ref{alg:hbn} shows one training iteration with HBN.

Note that HBN \emph{can not} be applied to SS-SS algorithms that do not use a teacher-student framework with EMA updates.

\begin{figure}[t]
  \centering
  \includegraphics[width=0.90\linewidth]{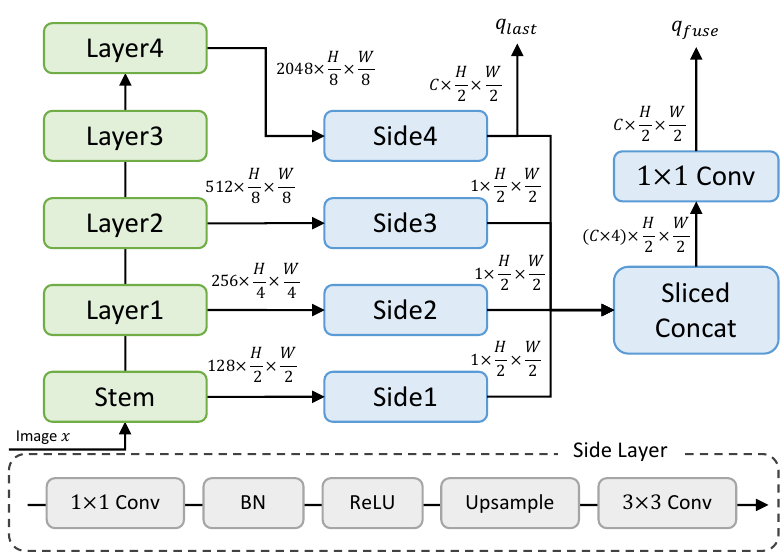}
  \caption{Architecture of the boundary detection head. The boundary head consists of four ``Side Layers'' which consists of a $1\times 1$ convolution up sampling to $\nicefrac{1}{2}$ of the input image size, and a $3\times 3$ convolution. The outputs are then fused together with a sliced concatenation operation followed by a $1\times 1$ convolution to produce the final boundary prediction.}
  \label{fig:app_boundary_head}
\vspace{-5pt} 
\end{figure}

\begin{figure}[t]
  \centering
  \includegraphics[width=0.90\linewidth]{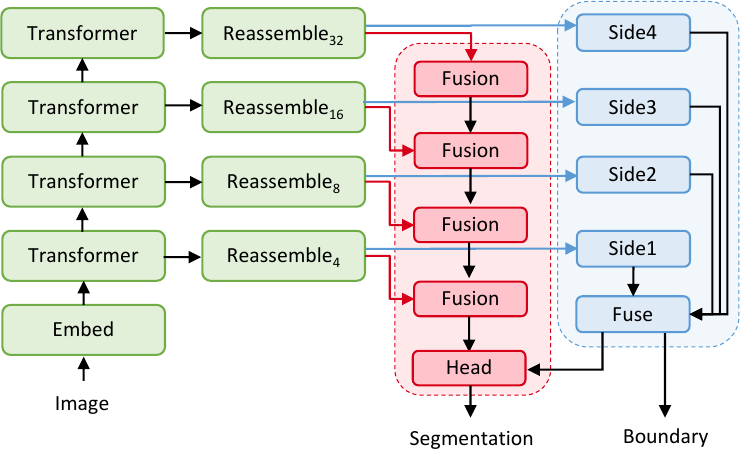}
  \caption{Architecture of DPT with boundary detection head used for BoundMatch framework.}
  \label{fig:app_dpt_architecture}
\vspace{-5pt} 
\end{figure}

\section{Architecture Details}

\subsection{Boundary Detection Head Design}
\label{app:boundary_head}

In \cref{fig:app_boundary_head}, we illustrate the architecture of the boundary detection head used in our BoundMatch framework.
We follow CASENet \cite{yu2017casenet} and use a multi-scale boundary detection head that consists of four ``Side Layers''.
Each ``Side Layer'' consists of a $1\times 1$ convolution up sampling to $\nicefrac{1}{2}$ of the input image size, followed by a $3\times 3$ convolution similar to the one introduced in SBCB \cite{ishikawa2023SBCB}.
For more details on the boundary detection head, please refer to the original paper \cite{yu2017casenet}.

\subsection{DINOv2 with BoundMatch}
\label{app:boundmatch_dpt}

In \cref{fig:app_dpt_architecture}, we illustrate the architecture of DPT with boundary detection head used in our BoundMatch framework.
Following the design of DPT, we use the reassembled feature maps used for the Fusion modules as the hierarchical features for the boundary detection head.
For more details on DPT, please refer to the original paper \cite{Ranftl2021DPT}.

For training under semi-supervised setting with BoundMatch, we use the same hyperparameters introduced in \cite{yang2025unimatchv2}.
Other hyperparameters such as $\lambda_{bdry}$ and $\tau_{bdry}$ remain the same as those used in SAMTH.

\subsection{Lightweight Models with BoundMatch}
\label{app:lightweight_models}

For DeepLabV3+ with MobileNet-V2 \cite{Sandler2018MobileNetV2}, we attach the boundary detection head to the output of the backbone in the same manner as DeepLabV3+ with ResNet backbones.
The hyperparameters used for training are identical to the setup introduced in \cref{sec:experimental_configuration}.

For integrating BoundMatch to AFFormer-Tiny \cite{Dong2023AFFormer}, we follow Mobile-Seed \cite{liao2024mobileseed} and use four features from the backbone (Blocks 1, 3, 5, and 6).
The input to the decoder assumes input channel size of 216 and the feature channel size before the classification layer is 96, which is identical to Mobile-Seed and AFFormer.
For feature fusion with BSF, we use one convolutional layer which takes in concatenated features from the backbone and boundary detection head to obtain the final feature.
The training hyperparamters are identical to the one used in AFFormer \cite{Dong2023AFFormer}.

\section{Boundary Region Evaluation}

\subsection{Boundary Evaluation Metrics}
\label{app:boundary_metrics}

\paragraph{Boundary IoU (BIoU):}
BIoU evaluates the overlap between predicted ($\hat{y}$) and ground‑truth segmentation masks $y$ restricted only to pixels lying within a narrow-band around object boundaries.
Formally, BIoU is computed as
\begin{equation}
\mathrm{BIoU} \;=\;\frac{1}{C}\sum_{c=1}^C 
\frac{\sum_{n}\sum_{(i,j)\in BD_k(y_n)} \bigl[\hat y_{n,c,i,j}\,\wedge\,y_{n,c,i,j}\bigr]}
{\sum_{n}\sum_{(i,j)\in BD_k(y_n)} \bigl[\hat y_{n,c,i,j}\,\vee\,y_{n,c,i,j}\bigr]},
\end{equation}
where \(BD_k(y) = y \oplus \mathrm{MinPool}_k(y)\) denotes the binary boundary mask obtained by applying a $k\times k$ min‑pooling (stride 1) to the ground‑truth segmentation map \(y\) and taking the pixel‑wise XOR (\(\oplus\)) with the original mask, and \(\wedge\) and \(\vee\) denote pixel‑wise logical AND and OR, respectively \cite{Kohli2008RobustHO}.
Because it ignores interior pixels, BIoU directly measures boundary alignment and penalizes both over‑ and under‑segmentation along object edges.
It is therefore particularly sensitive to boundary precision but agnostic to interior region accuracy.
We use a 5 pixel radius ($k=11$) for the boundary mask in our experiments.

\paragraph{Boundary F1 Score (BF1):}
In contrast, BF1 captures both boundary precision (the fraction of predicted boundary pixels that lie near a true boundary) and recall (the fraction of true boundary pixels recovered by the prediction), combining them via the harmonic mean:
\begin{equation}
\mathrm{BF1} \;=\;\frac{1}{C}\sum_{c=1}^C 
\frac{2\,\mathrm{Prec}_c\,\mathrm{Rec}_c}{\mathrm{Prec}_c + \mathrm{Rec}_c},
\end{equation}
where precision and recall are computed by matching predicted and ground‑truth boundary pixels within a small tolerance radius \cite{Csurka2013WhatIA}.
Formally, precision and recall can be defined as:
\footnotesize
\begin{equation}
\mathrm{Prec}_c \;=\;
\frac{
\sum_{n,i,j} \bigl[\mathrm{BD}_k(\hat y_n)_{c,i,j} \,\wedge\, \mathrm{MaxPool}_k\bigl(\mathrm{BD}_k(y_n)\bigr)_{c,i,j}\bigr]
}{
\sum_{n,i,j} \mathrm{BD}_k(\hat y_n)_{c,i,j}
},
\end{equation}
\normalsize
and
\footnotesize
\begin{equation}
\mathrm{Rec}_c \;=\;
\frac{
\sum_{n,i,j} \bigl[\mathrm{MaxPool}_k\bigl(\mathrm{BD}_k(\hat y_n)\bigr)_{c,i,j} \,\wedge\, \mathrm{BD}_k(y_n)_{c,i,j}\bigr]
}{
\sum_{n,i,j} \mathrm{BD}_k(y_n)_{c,i,j}
}.
\end{equation}
\normalsize

BF1 thus provides insight into the trade‑off between missing fine boundary details (low recall) and producing spurious edges (low precision), while allowing for minor spatial misalignments.
We use a 5 pixel tolerance radius ($k=11$) for our experiments.

\paragraph{Complementary Insights:}
BIoU emphasizes exact boundary overlap, making it well suited for tasks requiring crisp, well‑aligned boundaries.
BF1, by explicitly balancing precision and recall under a spatial tolerance, is more robust to slight shifts but sensitive to fragmented or noisy boundary predictions.
Reporting both metrics alongside mIoU therefore provides a holistic view of a model’s segmentation performance, particularly its ability to delineate object boundaries accurately.

\subsection{BIoU and BF1 Evaluation of Benchmark Results}
\label{app:boundary_metrics_results}

\begin{table}[t]
\centering
\caption{Component analysis of the proposed modules on Pascal VOC 2012 val set using ResNet-50 backbone. We report the mean IoU, Boundary IoU (BIoU), and Boundary F1 (BF1) scores. The best results are highlighted in bold.}
\label{tab:ablation_components_voc}
\footnotesize
\setlength{\tabcolsep}{3pt}
\begin{tabular}{
  c c c
  S[table-format=2.1] S[table-format=2.1] S[table-format=2.1] 
  S[table-format=2.1] S[table-format=2.1] S[table-format=2.1] 
}
\toprule
\multicolumn{3}{c}{} &
\multicolumn{3}{c}{$\nicefrac{1}{16}$} &
\multicolumn{3}{c}{$\nicefrac{1}{8}$} \\
\cmidrule(lr){4-6}\cmidrule(lr){7-9}
\textbf{BCRM} & \textbf{BSF} & \textbf{SGF} &
\multicolumn{1}{c}{IoU} & \multicolumn{1}{c}{BIoU} & \multicolumn{1}{c}{BF1} &
\multicolumn{1}{c}{IoU} & \multicolumn{1}{c}{BIoU} & \multicolumn{1}{c}{BF1} \\
\midrule
--     & --     & --     &
73.5 & 67.9 & 59.0 &
75.8 & 68.7 & 59.2 \\
\cmark & --     & --     &
75.6 & 68.3 & 59.2 &
76.8 & 69.1 & 59.3 \\
\cmark & \cmark & --     &
75.9 & 68.1 & 59.3 &
76.5 & 69.3 & 60.0 \\
\cmark & --     & \cmark &
75.8 & 68.0 & 59.2 &
76.6 & 68.8 & 59.8 \\
\rowcolor{ablrow}
\cmark & \cmark & \cmark &
\textbf{76.3} & \textbf{68.9} & \textbf{59.8} &
\textbf{77.2} & \textbf{69.7} & \textbf{60.3} \\
\bottomrule
\end{tabular}
\vspace{-10pt} 
\end{table}

\begin{table*}[t] 
  \centering
  \setlength{\tabcolsep}{3.5pt} 
  \begin{minipage}{.48\textwidth}\centering
    \captionof{table}{Boundary Metrics for Cityscapes Benchmark.}
    \label{tab:cityscapes_boundary_metrics}
    \scriptsize
    \setlength{\tabcolsep}{3pt}
    \begin{tabular}{
      l
      S[table-format=2.1] S[table-format=2.1]  
      S[table-format=2.1] S[table-format=2.1]  
      S[table-format=2.1] S[table-format=2.1]  
      S[table-format=2.1] S[table-format=2.1]  
    }
    \toprule
    & \multicolumn{4}{c}{\textbf{ResNet-50}} & \multicolumn{4}{c}{\textbf{ResNet-101}} \\
    \cmidrule(lr){2-5}\cmidrule(l){6-9}
    & \multicolumn{2}{c}{$\nicefrac{1}{16}$} & \multicolumn{2}{c}{$\nicefrac{1}{8}$}
    & \multicolumn{2}{c}{$\nicefrac{1}{16}$} & \multicolumn{2}{c}{$\nicefrac{1}{8}$} \\
    \cmidrule(lr){2-3}\cmidrule(lr){4-5}\cmidrule(lr){6-7}\cmidrule(l){8-9}
    & \multicolumn{1}{c}{BIoU} & \multicolumn{1}{c}{BF1}
    & \multicolumn{1}{c}{BIoU} & \multicolumn{1}{c}{BF1}
    & \multicolumn{1}{c}{BIoU} & \multicolumn{1}{c}{BF1}
    & \multicolumn{1}{c}{BIoU} & \multicolumn{1}{c}{BF1} \\
    \midrule
    SAMTH
    & 54.8 & 59.3
    & 56.8 & 61.8
    & 55.8 & 61.4
    & 57.9 & 63.1 \\
    \quad + BoundMatch
    & 56.6 & 63.2
    & 57.8 & 64.7
    & 58.3 & 65.2
    & 59.4 & 66.7 \\
    \midrule
    UniMatch
    & 55.3 & 60.7
    & 56.6 & 63.5
    & 56.9 & 63.3
    & 57.8 & 65.1 \\
    \quad + BoundMatch
    & 56.0 & 62.6
    & 57.5 & 64.2
    & 57.6 & 64.8
    & 58.8 & 66.4 \\
    \midrule
    PrevMatch
    & 55.3 & 60.7
    & 57.0 & 63.3
    & 57.5 & 63.7
    & 58.8 & 65.6 \\
    \quad + BoundMatch
    & 56.8 & 63.4
    & 58.2 & 64.4
    & 58.0 & 65.4
    & 58.9 & 66.4 \\
    \bottomrule
    \end{tabular}
  \end{minipage}\hfill
  \begin{minipage}{.48\textwidth}\centering
    \captionof{table}{Boundary Metrics for Pascal VOC Classic Split.}
    \label{tab:boundary_voc_classic}
    \scriptsize
    \setlength{\tabcolsep}{3pt}
    \begin{tabular}{
      l
      S[table-format=2.1] S[table-format=2.1]  
      S[table-format=2.1] S[table-format=2.1]  
      S[table-format=2.1] S[table-format=2.1]  
      S[table-format=2.1] S[table-format=2.1]  
    }
    \toprule
    & \multicolumn{4}{c}{\textbf{ResNet-50}} & \multicolumn{4}{c}{\textbf{ResNet-101}} \\
    \cmidrule(lr){2-5}\cmidrule(l){6-9}
    & \multicolumn{2}{c}{92} & \multicolumn{2}{c}{183}
    & \multicolumn{2}{c}{92} & \multicolumn{2}{c}{183} \\
    \cmidrule(lr){2-3}\cmidrule(lr){4-5}\cmidrule(lr){6-7}\cmidrule(l){8-9}
    & \multicolumn{1}{c}{BIoU} & \multicolumn{1}{c}{BF1}
    & \multicolumn{1}{c}{BIoU} & \multicolumn{1}{c}{BF1}
    & \multicolumn{1}{c}{BIoU} & \multicolumn{1}{c}{BF1}
    & \multicolumn{1}{c}{BIoU} & \multicolumn{1}{c}{BF1} \\
    \midrule
    SAMTH
    & 65.2 & 55.0
    & 65.6 & 53.4
    & 70.7 & 59.5
    & 70.5 & 59.6 \\
    \quad + BoundMatch
    & 67.2 & 56.9
    & 68.3 & 57.6
    & 70.9 & 61.3
    & 72.7 & 62.5 \\
    \midrule
    UniMatch
    & 64.9 & 55.2
    & 65.2 & 56.1
    & 69.3 & 57.4
    & 69.9 & 60.1 \\
    \quad + BoundMatch
    & 68.2 & 57.7
    & 68.6 & 57.5
    & 70.2 & 59.0
    & 72.1 & 61.7 \\
    \midrule
    PrevMatch
    & 67.1 & 57.1
    & 67.3 & 57.2
    & 69.8 & 60.7
    & 71.8 & 61.8 \\
    \quad + BoundMatch
    & 69.0 & 58.0
    & 68.7 & 57.4
    & 71.0 & 61.2
    & 72.4 & 61.8 \\
    \bottomrule
    \end{tabular}
  \end{minipage}

  \vspace{6pt} 

  \begin{minipage}{.42\textwidth}\centering
    \captionof{table}{Boundary Metrics for Pascal VOC Blender Split.}
    \label{tab:boundary_voc_blender}
    \scriptsize
    \setlength{\tabcolsep}{3pt}
    \begin{tabular}{
      l
      S[table-format=2.1] S[table-format=2.1]  
      S[table-format=2.1] S[table-format=2.1]  
    }
    \toprule
    & \multicolumn{2}{c}{$\nicefrac{1}{16}$} & \multicolumn{2}{c}{$\nicefrac{1}{8}$} \\
    \cmidrule(lr){2-3}\cmidrule(lr){4-5}
    & \multicolumn{1}{c}{BIoU} & \multicolumn{1}{c}{BF1}
    & \multicolumn{1}{c}{BIoU} & \multicolumn{1}{c}{BF1} \\
    \midrule
    SAMTH
    & 68.5 & 59.8
    & 69.2 & 59.1 \\
    \quad + BoundMatch
    & 68.8 & 60.1
    & 70.0 & 59.9 \\
    \midrule
    UniMatch
    & 66.6 & 57.3
    & 68.0 & 57.2 \\
    \quad + BoundMatch
    & 69.4 & 59.5
    & 70.4 & 60.4 \\
    \midrule
    PrevMatch
    & 68.4 & 57.7
    & 69.3 & 58.2 \\
    \quad + BoundMatch
    & 68.7 & 59.1
    & 69.4 & 59.3 \\
    \bottomrule
    \end{tabular}
  \end{minipage}\hfill
  \begin{minipage}{.57\textwidth}\centering
    \captionof{table}{Boundary Metrics for BDD100K, SYNTHIA, and ADE20K.}
    \label{tab:boundary_other}
    \scriptsize
    \setlength{\tabcolsep}{2pt}
    \begin{tabular}{
      l
      S[table-format=2.1] S[table-format=2.1]  
      S[table-format=2.1] S[table-format=2.1]  
      S[table-format=2.1] S[table-format=2.1]  
      S[table-format=2.1] S[table-format=2.1]  
      S[table-format=2.1] S[table-format=2.1]  
      S[table-format=2.1] S[table-format=2.1]  
    }
    \toprule
    & \multicolumn{4}{c}{\textbf{BDD100K}} & \multicolumn{4}{c}{\textbf{SYNTHIA}} & \multicolumn{4}{c}{\textbf{ADE20K}} \\
    \cmidrule(lr){2-5}\cmidrule(l){6-9}\cmidrule(l){10-13}
    & \multicolumn{2}{c}{$\nicefrac{1}{64}$} & \multicolumn{2}{c}{$\nicefrac{1}{32}$}
    & \multicolumn{2}{c}{$\nicefrac{1}{64}$} & \multicolumn{2}{c}{$\nicefrac{1}{32}$}
    & \multicolumn{2}{c}{$\nicefrac{1}{128}$} & \multicolumn{2}{c}{$\nicefrac{1}{64}$} \\
    \cmidrule(lr){2-3}\cmidrule(lr){4-5}
    \cmidrule(lr){6-7}\cmidrule(lr){8-9}
    \cmidrule(lr){10-11}\cmidrule(lr){12-13}
    & \multicolumn{1}{c}{BIoU} & \multicolumn{1}{c}{BF1}
    & \multicolumn{1}{c}{BIoU} & \multicolumn{1}{c}{BF1}
    & \multicolumn{1}{c}{BIoU} & \multicolumn{1}{c}{BF1}
    & \multicolumn{1}{c}{BIoU} & \multicolumn{1}{c}{BF1}
    & \multicolumn{1}{c}{BIoU} & \multicolumn{1}{c}{BF1}
    & \multicolumn{1}{c}{BIoU} & \multicolumn{1}{c}{BF1} \\
    \midrule
    Supervised
    & 31.8 & 34.2 & 35.9 & 37.4
    & 55.3 & 68.7 & 60.3 & 74.8
    & 5.8 & 6.9 & 7.6 & 8.3 \\
    UniMatch
    & 37.9 & 42.5 & 41.1 & 45.5
    & 60.8 & 74.5 & 63.6 & 78.1
    & 9.2 & 10.2 & 12.7 & 12.8 \\
    Ours
    & 39.8 & 46.1 & 41.3 & 47.3
    & 62.2 & 76.7 & 64.8 & 80.1
    & 12.0 & 13.0 & 14.9 & 16.2 \\
    \bottomrule
    \end{tabular}
  \end{minipage}
  \vspace{-10pt} 
\end{table*}

In \cref{tab:cityscapes_boundary_metrics,tab:boundary_voc_classic,tab:boundary_voc_blender,tab:boundary_other}, we show the boundary evaluation metrics (BIoU and BF1) for each of the datasets used in the main paper.
While comprehensive boundary evaluation of all prior methods would be ideal, most existing SS-SS works do not provide open-source implementations or pretrained models, making it infeasible to fairly compute these specialized metrics retrospectively.
We therefore present boundary metrics for the representative methods we could reliably reproduce (SAMTH, UniMatch, PrevMatch) alongside their BoundMatch-enhanced versions.
Despite this limitation, the results clearly demonstrate that BoundMatch consistently improves boundary quality across different datasets and baseline models, validating our approach's effectiveness in enhancing boundary delineation.

\section{Additional Ablation Studies}

\subsection{Component Analysis of Pascal VOC}
\label{app:ablation_components_voc}

In \cref{tab:ablation_components_voc}, we show the component analysis of our BoundMatch framework on Pascal VOC Classic 92 with ResNet-50 backbone.
We can see that each component of BoundMatch contributes to the overall performance.
The most contribution comes from the Boundary-aware Consistency Regularization Module (BCRM), which improves over 1.0\% compared to the SAMTH baseline.
Incremental, but gradual improvements can be seen when introducing BSF and SGF, and results in over 1.4--2.8\% improvements.

\subsection{Pseudo-Label Accuracy}
\label{app:pseudo_label_accuracy}

\begin{figure}[t] 
  \centering
  \subfloat[Cityscapes $\nicefrac{1}{16}$.\label{fig:a}]{%
    \includegraphics[width=0.49\linewidth]{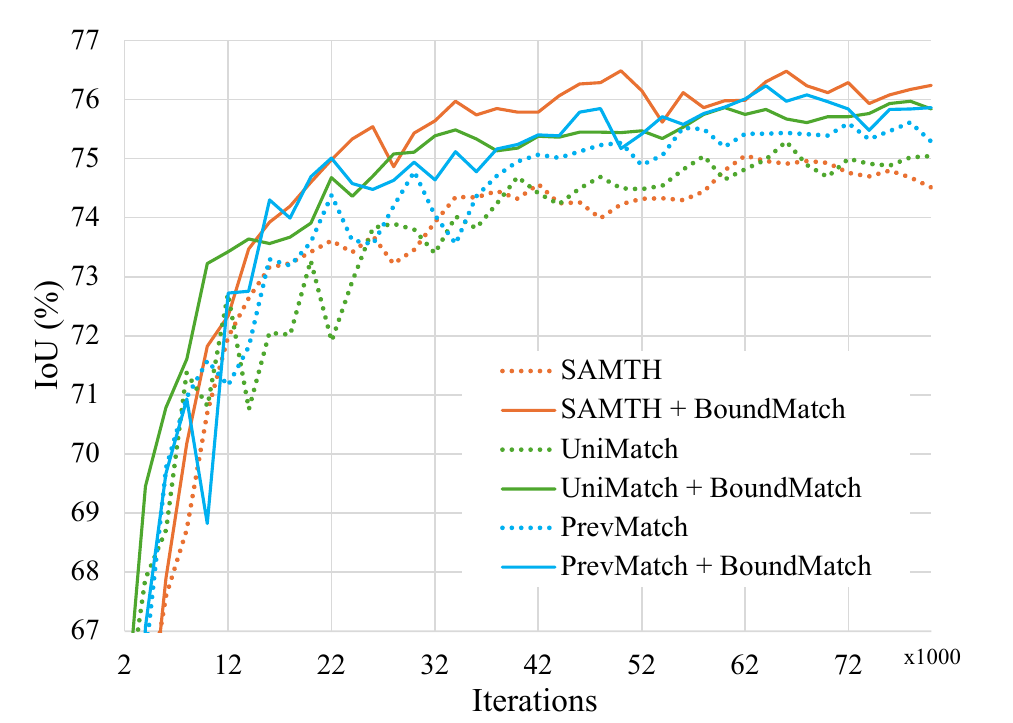}%
  }\hfill
  \subfloat[Pascal VOC Classic 92.\label{fig:b}]{%
    \includegraphics[width=0.49\linewidth]{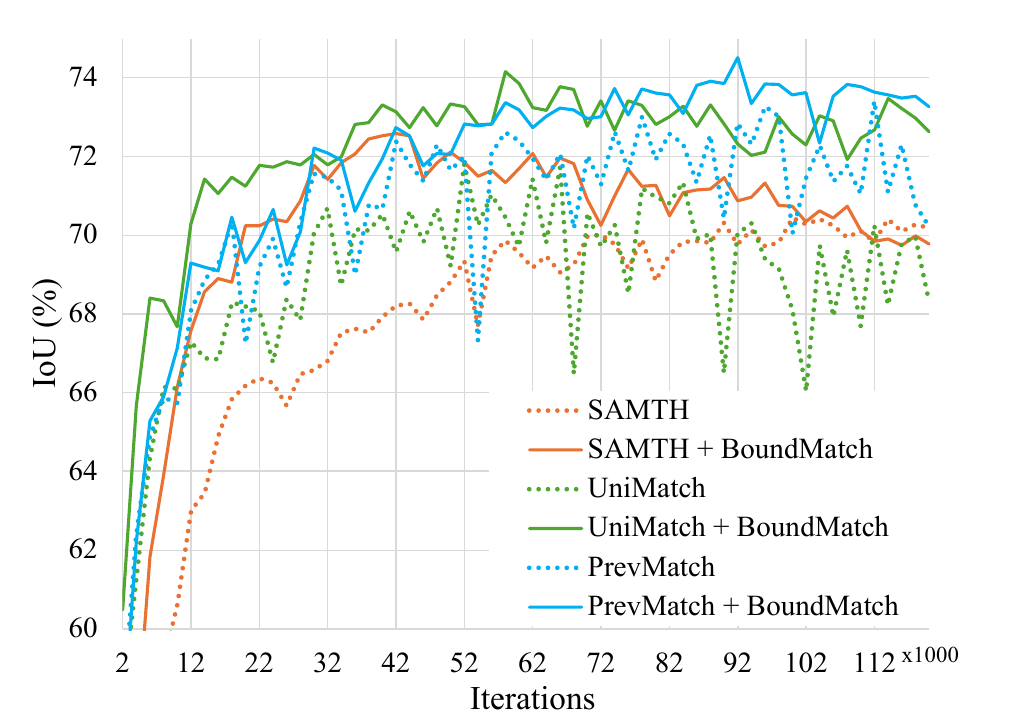}%
  }
  \caption{(a) and (b) presents pseudo-label accuracies (mIoU) of the validation set against training iterations for Cityscapes and Pascal VOC Classic 92, respectively. We compare SAMTH, UniMatch, and PrevMatch with and without BoundMatch.}
  \label{fig:pseudo_label_accuracy}
\vspace{-10pt} 
\end{figure}

In \cref{fig:pseudo_label_accuracy}, we plot the pseudo-label accuracy (IoU) on the validation set against training iterations for Cityscapes and Pascal VOC Classic 92.
The training curves reveal BoundMatch's noise-robustness properties.
On Cityscapes \cref{fig:a}, all methods (with and without BoundMatch) show monotonically increasing pseudo-label accuracy without late-stage degradation, indicating that consistency regularization prevents overfitting to early noisy pseudo-labels.
On Pascal VOC \cref{fig:b}, where boundary annotations are often noisier, BoundMatch notably stabilizes the fluctuating pseudo-label accuracy observed in baseline methods, particularly for UniMatch and PrevMatch.
SAMTH+BoundMatch shows improvements over SAMTH in terms of accuracy, but overfitting can be seen after 40K iterations, indicating the need for stronger regularizations which are equipped in UniMatch and PrevMatch.

This stabilization can be attributed to the multi-task learning framework: when segmentation pseudo-labels become noisy, the boundary task, learned from hierarchical features and refined through SGF, provides a strong regularization signal to deter the model in overfitting.
This aligns with the noise-robust learning principles identified in \cite{Zhang2024NRCR}, where combining multiple complementary views and regularization mechanisms proves more effective than single approaches.

\section{Additional Qualitative Results}

\begin{figure*}[t]
  \centering
  \subfloat[Improved Cases.\label{fig:app_successful_cases}]{%
    \includegraphics[width=0.48\linewidth]{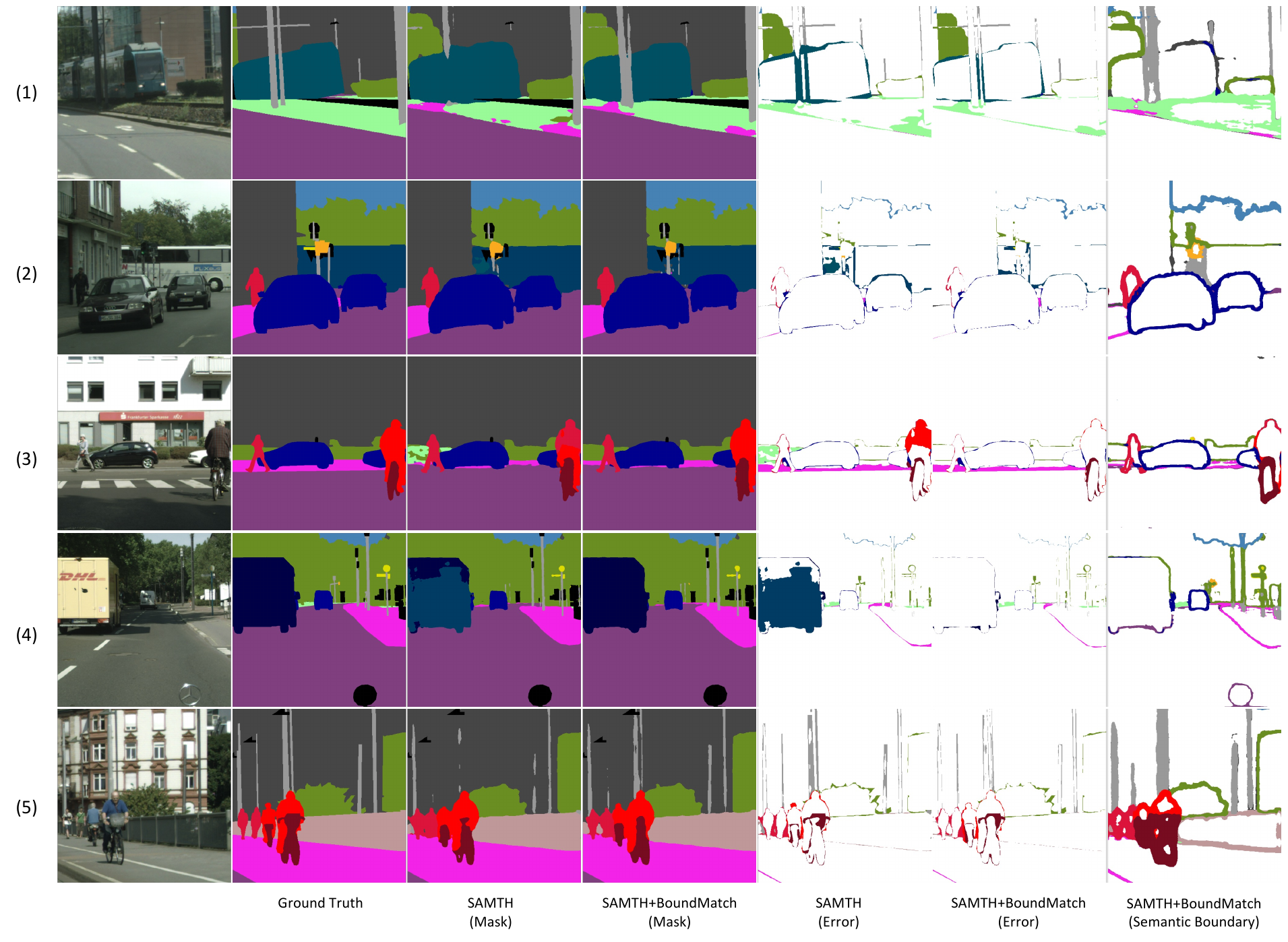}%
  }\hfill
  \subfloat[Failure Cases.\label{fig:app_failure_cases}]{%
    \includegraphics[width=0.48\linewidth]{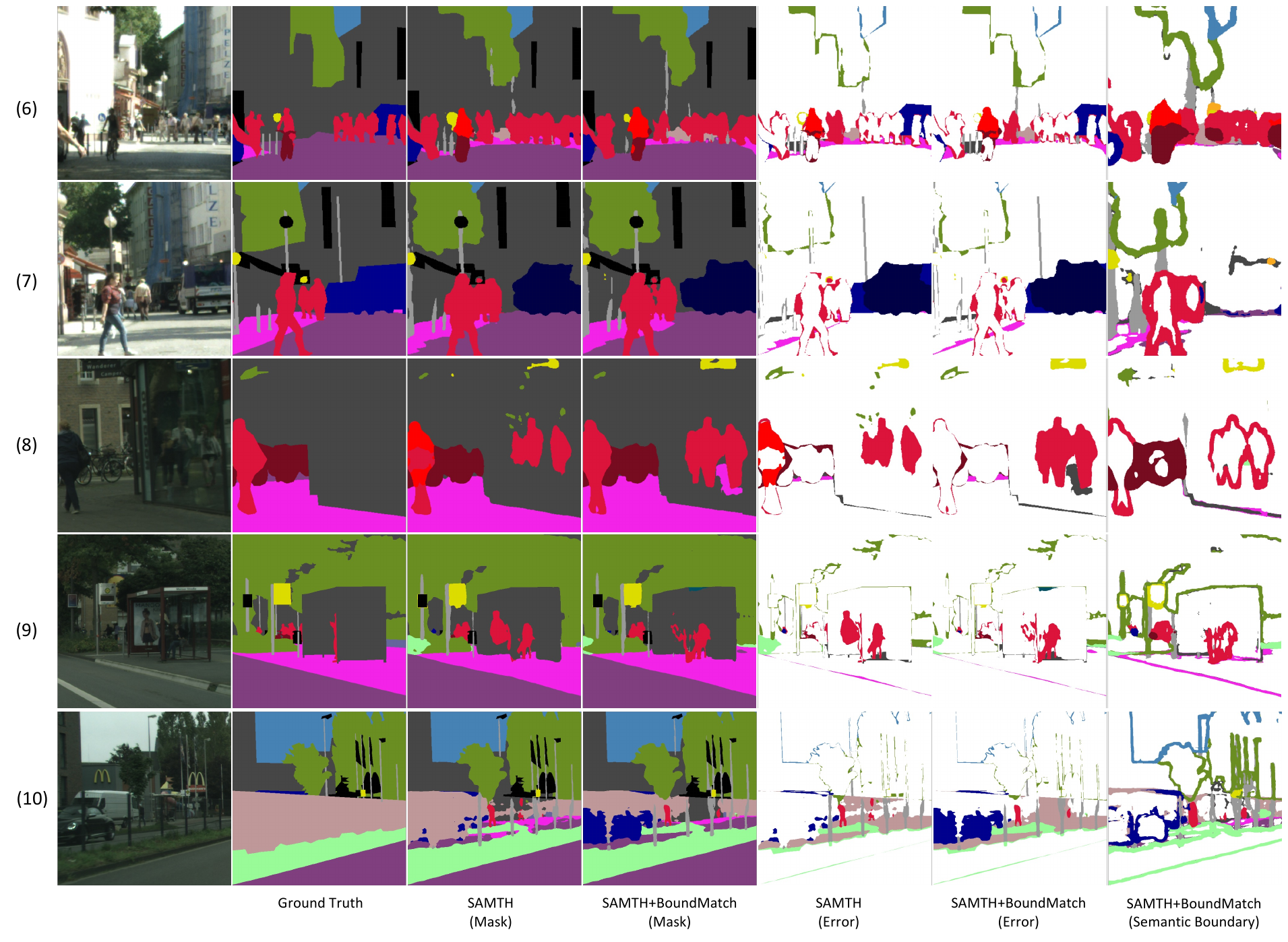}%
  }
  \vspace{-0pt}
  \caption{More qualitative results which compare SAMTH and SAMTH+BoundMatch on the Cityscapes dataset with $\nicefrac{1}{16}$ protocol: (a) improved cases and (b) failure cases. Best viewed zoomed in and in color. Note that we cropped and scaled each samples differently across each samples and boundary predictions could seem thicker due to being zoomed in.}
  \vspace{-5pt}
\end{figure*}

\begin{figure}[t]
  \centering
  \includegraphics[width=\linewidth]{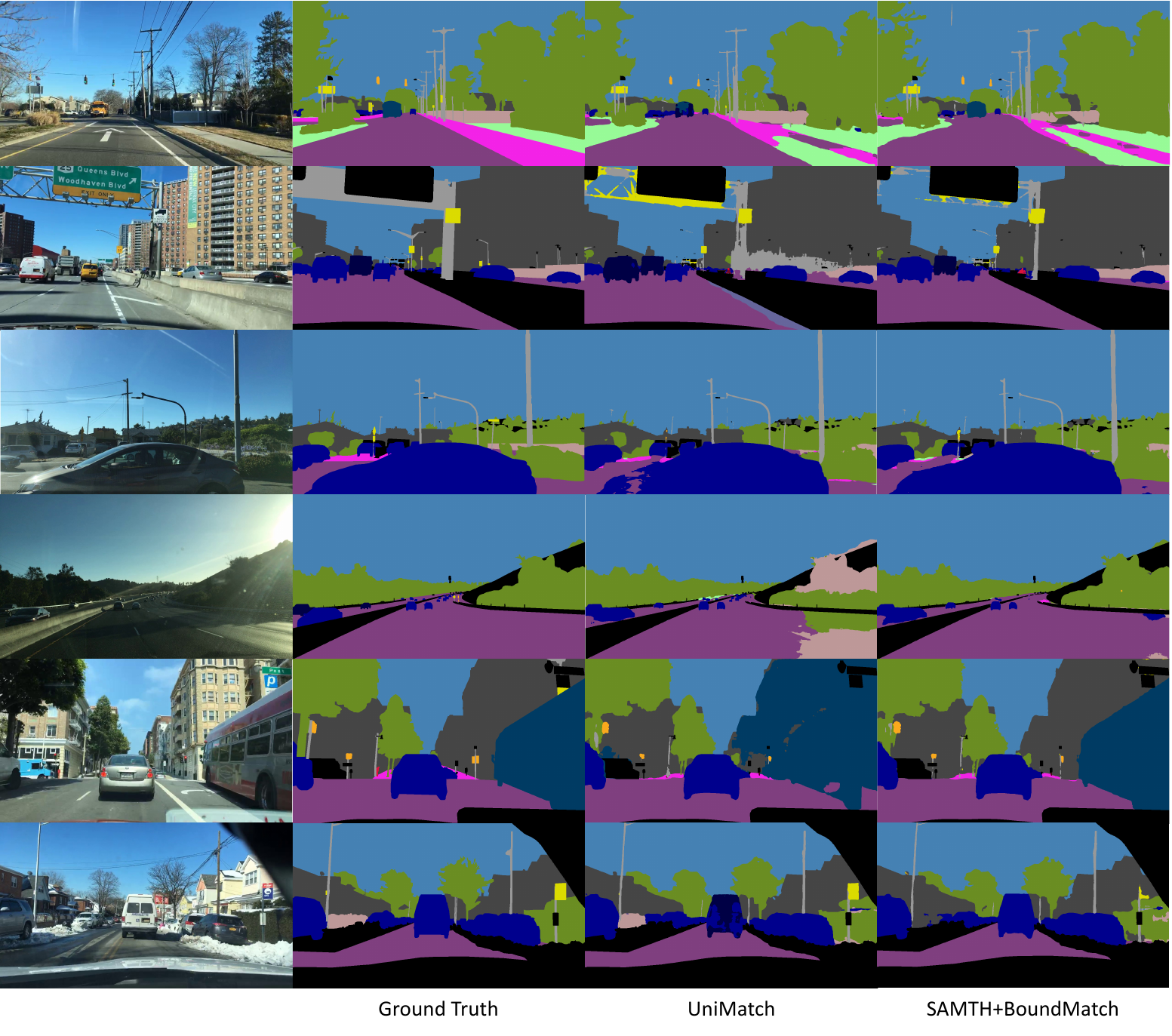}
  \vspace{-15pt} 
  \caption{Qualitative results on the BDD100K dataset on $\nicefrac{1}{64}$ split. We compare UniMatch \cite{Yang2022UniMatch} with our SAMTH+BoundMatch. Best viewed zoomed in and in color.}
  \label{fig:app_bdd100k_qualitative}
\vspace{-5pt} 
\end{figure}

\begin{figure}[t]
  \centering
  \includegraphics[width=\linewidth]{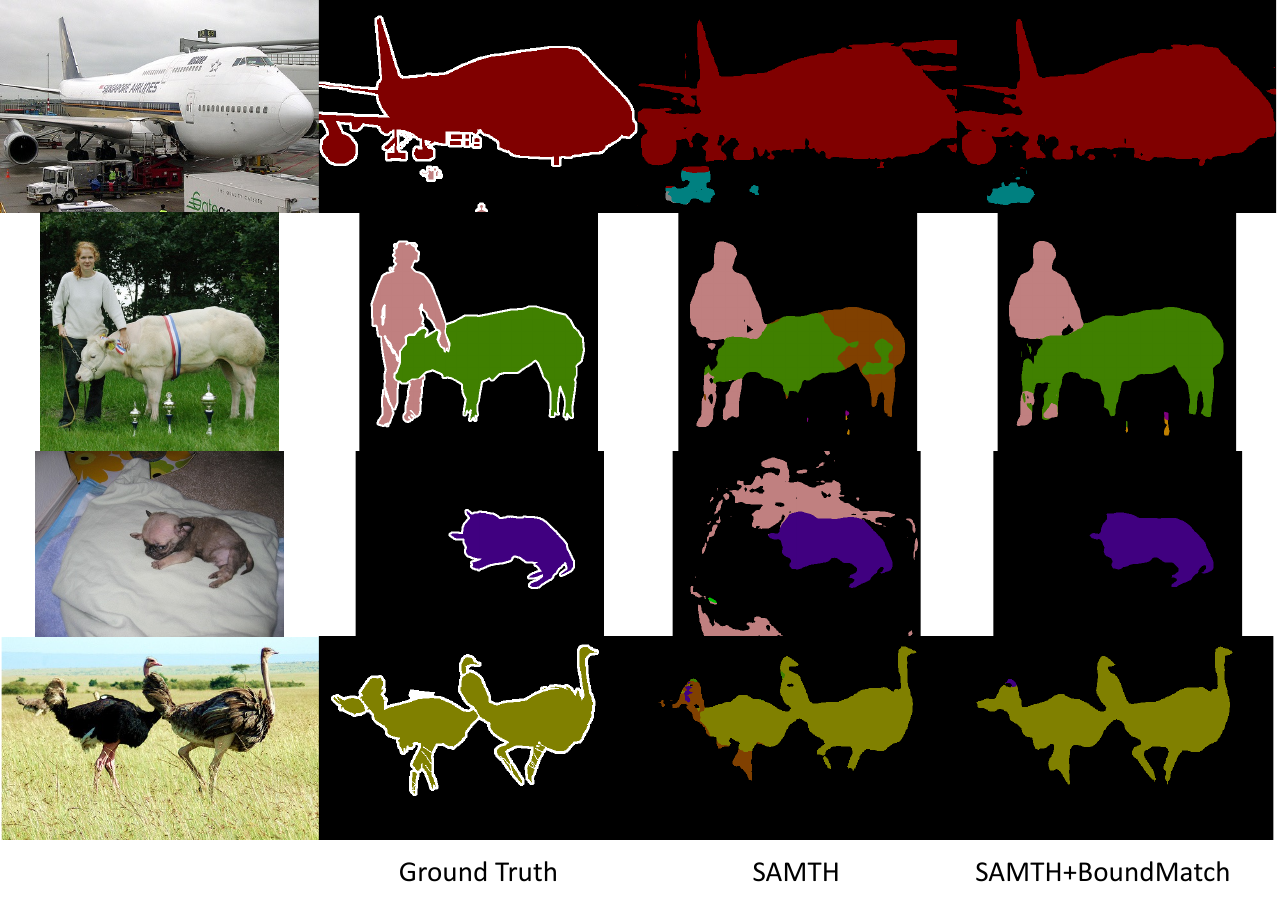}
 \vspace{-15pt} 
  \caption{Qualitative results on the Pascal VOC 2012 dataset on the difficult \textit{Classic} 92 image split. We compare SAMTH and SAMTH+BoundMatch. The white region in the ground truth masks are ``ignore'' regions. Best viewed zoomed in and in color.}
  \label{fig:app_voc_qualitative}
\vspace{-5pt} 
\end{figure}

\begin{figure}[t]
  \centering
  \includegraphics[width=\linewidth]{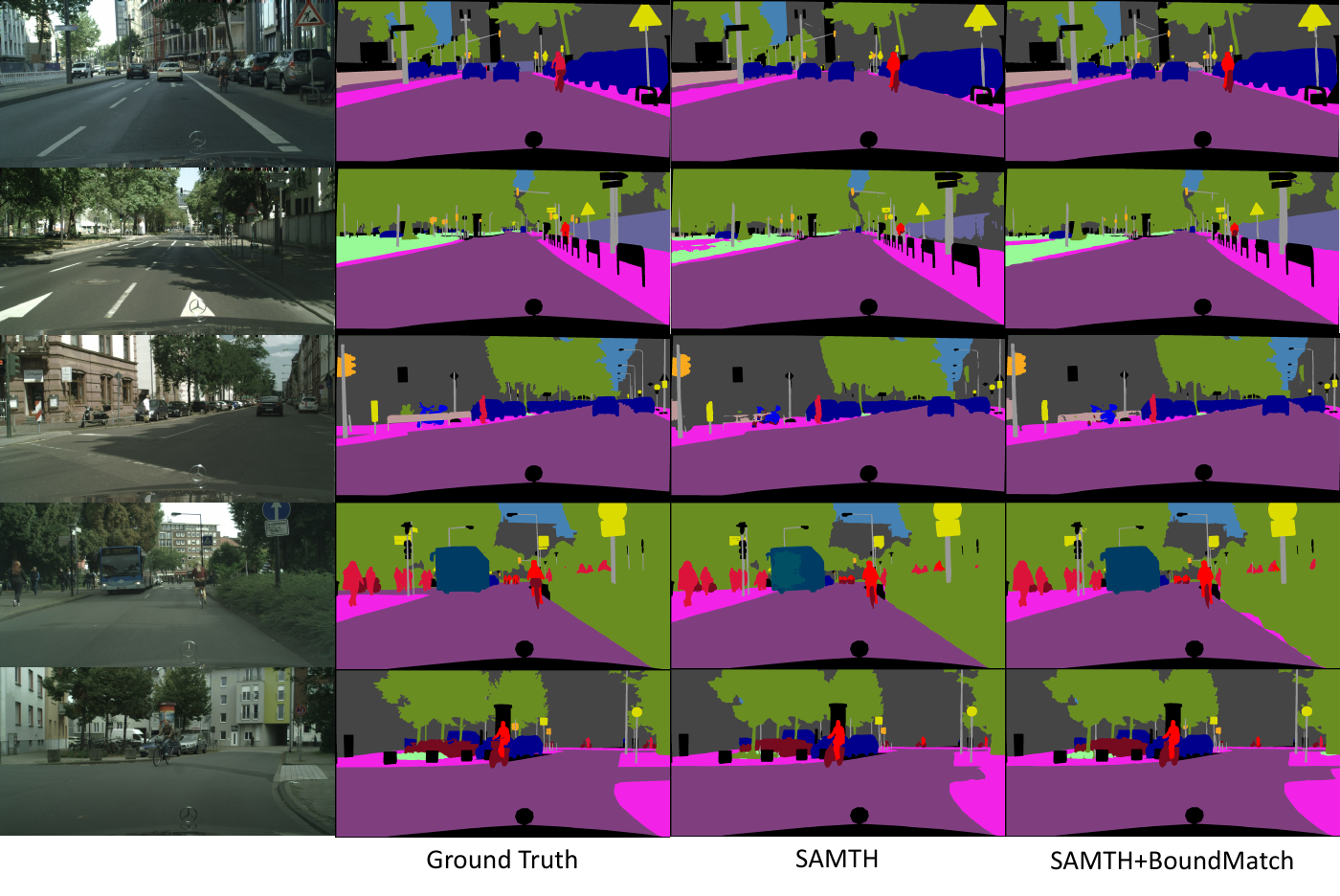}
  \vspace{-15pt} 
  \caption{Qualitative results on the Cityscapes dataset on $\nicefrac{1}{16}$ split. We compare SAMTH and SAMTH+BoundMatch using DPT with \textbf{DINOv2-S} pretrained ViT encoder. Best viewed zoomed in and in color.}
  \label{fig:app_dinvo2_small_qualitative}
\vspace{-5pt} 
\end{figure}

\begin{table}[t]
\centering
\caption{Results on the ACDC Dataset. The results are reported in Dice Similarity Coefficient.}
\label{tab:benchmark_acdc}
\footnotesize
\begin{tabular}{
  l
  S[table-format=2.1]
  S[table-format=2.1]
  S[table-format=2.1]
}
\toprule
& \multicolumn{3}{c}{\textbf{Cases}}\\
\cmidrule(l){2-4}
& 1 & 3 & 7 \\
\midrule
Supervised & 28.5 & 41.5 & 62.5\\
CPS \cite{Chen2021CPS} & \multicolumn{1}{c}{—} & 61.0 & 81.5\\
UniMatch \cite{Yang2022UniMatch} & 85.4 & 88.9 & 89.9\\
\rowcolor{ablrow}
\textbf{UniMatch + BoundMatch} & 85.0 & 89.0 & 89.6\\
\bottomrule
\end{tabular}
\vspace{-10pt} 
\end{table}

\begin{table}[t]
\centering
\caption{Results on the LoveDA dataset.}
\label{tab:benchmark_loveda}
\scriptsize
\setlength{\tabcolsep}{3pt}
\begin{tabular}{
  l
  S[table-format=2.1] S[table-format=2.1] S[table-format=2.1]
  S[table-format=2.1] S[table-format=2.1] S[table-format=2.1]
}
\toprule
& \multicolumn{3}{c}{\nicefrac{1}{32}} & \multicolumn{3}{c}{\nicefrac{1}{16}}\\
\cmidrule(l){2-4}\cmidrule(l){5-7}
& IoU & BIoU & {BF1} & IoU & BIoU & {BF1}\\
\midrule
Supervised & 45.2 & 27.0 & 24.8 & 45.7 & 29.1 & 25.3\\
UniMatch & 48.9 & 28.7 & 28.3 & 48.8 & 29.4 & 29.7\\
SAMTH & 49.8 & 29.7 & 29.8 & 50.1 & 30.7 & 29.6\\
\rowcolor{ablrow}
\textbf{SAMTH + BoundMatch} & 51.1 & 30.3 & 30.8 & 52.1 & 31.3 & 31.1\\
\bottomrule
\end{tabular}
\vspace{-10pt} 
\end{table}

\subsection{Cityscapes}
\label{app:cityscapes_qualitative}

In \cref{fig:app_successful_cases}, we present qualitative comparisons between SAMTH and SAMTH+BoundMatch on the Cityscapes dataset.
For each example, we display the segmentation masks, prediction errors, and semantic boundary predictions (for SAMTH+BoundMatch only).
BoundMatch reduces segmentation errors near object boundaries, as demonstrated in examples (1) through (5).
Notably, BoundMatch better preserves object continuity: in (3) and (4), it accurately segments complete "terrain", "rider", and "truck" regions rather than producing fragmented results.
Despite the inherent challenge of segmenting thin structures, BoundMatch successfully captures "poles" in examples (1) and (5), where the baseline typically struggles.
The strong alignment between predicted boundaries and actual object edges indicates that the segmentation head leverages boundary information to generate more accurate masks.

In \cref{fig:app_failure_cases}, we show notable failure cases for SAMTH+BoundMatch on the Cityscapes dataset.
We believe thin objects like "poles" show limited improvement for two main reasons: first, small objects create imbalanced supervision signals, and second, annotation inconsistencies lead to false positives.
For instance, images (6) and (7) show the same scene from different samples, but poles are annotated only in (7), not in (6).
This inconsistency causes false positives, particularly for SAMTH+BoundMatch, which aims to precisely segment boundaries.
BoundMatch also struggles with challenging scenarios like reflections, as seen in (8).
Additionally, transparent or semi-transparent structures pose difficulties: in (9) and (10), see-through "fences" and "buildings" challenge BoundMatch as it attempts to accurately segment partially occluded objects behind them.

\subsection{BDD100K}
\label{app:bdd100k_qualitative}
We show the qualitative results on the BDD100K dataset in \cref{fig:app_bdd100k_qualitative}.
Compared to UniMatch, our SAMTH+BoundMatch generally outputs more accurate segmentations.
For example, in the first and last rows, the bus and truck is accurately segmented with BoundMatch.
In the second and fifth rows, the road and buildings are accurately segmented with BoundMatch, where UniMatch fails to do so.
The fourth row also shows BoundMatch's robustness against sun flares and shadows, being able to segment the road more accurately.
However, there remains some challenges when the dataset's labels are ambiguous; such as the ``sidewalk'' region next to the road might be considered a ``vegetation'' in the first row.

\subsection{Pascal VOC 2012}
\label{app:voc_qualitative}
We show the qualitative results on the Pascal VOC 2012 dataset in \cref{fig:app_voc_qualitative}.
This dataset is especially challenging due to the presence of small objects and complex backgrounds, but our SAMTH+BoundMatch method demonstrates improved segmentation quality over SAMTH especially around object boundaries.
For example, the first row shows that BoundMatch is able to delineate the airplane and have better boundary quality compared to without using boundary consistency regularization.
Also, the third row also suggests that the model learns to separate the category of interest from the background more accurately.

\subsection{DINOv2-S}
\label{app:dino_qualitative}
We show the qualitative results on the Cityscapes dataset using DINOv2-S pretrained ViT in \cref{fig:app_dinvo2_small_qualitative}.
DINOv2-S is a strong foundation model that has been pre-trained on a large dataset, and it is known to perform well on various vision tasks.
As shown in the samples, BoundMatch is able to improve the segmentation quality compared to SAMTH.
For example, the first row shows that BoundMatch is able to delineate the road/sidewalk and the car around their boundaries more accurately.
Far away objects such as the wall and the thin poles are also better segmented with BoundMatch.
Other samples also provide similar observations.

\section{BoundMatch in other domains}
\label{app:other_domains}

In \cref{tab:benchmark_acdc}, we show the results on the ACDC dataset \cite{Bernard2018ACDC} which is a medical image segmentation dataset.
We follow the same experimental setup as in \cite{Yang2022UniMatch} and use UNet as the base model.
To integrate BoundMatch, we took the hierarchical features from the encoder to the boundary detection head follwing \cref{app:boundary_head}.
We found that BoundMatch does not improve the performance of UniMatch on this dataset.
This could be due to the fact that our architectures and losses for the introduced BoundMatch would need to be further optimized for medical images, which is out of the scope of this paper.

In \cref{tab:benchmark_loveda}, we show the results on the LoveDA dataset \cite{Wang2021LoveDA} which is a remote sensing image segmentation dataset.
We use DeepLabV3+ as the base model and produced results on $\nicefrac{1}{32}$ and $\nicefrac{1}{16}$ splits.
We can see that BoundMatch consistently improves the performance of SAMTH and UniMatch on this dataset, providing potentials that BoundMatch can be generalized to other domains.

\footnotesize
\bibliographystyle{unsrt}
\bibliography{refs}  

\end{document}